\documentclass{article}

\usepackage[numbers]{natbib}

\usepackage[preprint]{neurips_2025}

\usepackage[utf8]{inputenc} 
\usepackage[T1]{fontenc}    
\usepackage{hyperref}       
\usepackage{url}            
\usepackage{booktabs}       
\usepackage{amsfonts}       
\usepackage{nicefrac}       
\usepackage{microtype}      
\usepackage{xcolor}         
\usepackage{array}
\usepackage{amsmath}
\usepackage{siunitx}
\usepackage{glossaries}
\usepackage{graphicx}

\usepackage{ragged2e}

\usepackage{subcaption}
\usepackage{colortbl}
\definecolor{lightgreen}{RGB}{210, 240, 210}
\definecolor{lightred}{RGB}{240, 210, 210}
\definecolor{lightorange}{RGB}{252, 235, 210}

\newacronym{sota}{SOTA}{State-of-the-art}
\newacronym{lidar}{LiDAR}{LiDAR}
\newacronym{in-v2x}{IN-V2X}{Ingolstadt Vehicle-to-Everything}
\newacronym{fov}{FOV}{Field-of-View}
\newacronym{hfov}{HFOV}{Horizontal \gls{fov}}
\newacronym{v2x}{V2X}{Vehicle-to-Everything}
\newacronym{rtk}{RTK}{Real Time Kinematic}
\newacronym{fhd}{FHD}{Full High Definition}
\newacronym{fps}{FPS}{Frames Per Second}
\newacronym{ptp}{PTP}{Precision Time Protocol}
\newacronym{ntp}{NTP}{Network Time Protocol}
\newacronym{gps}{GPS}{Global Positioning System}
\newacronym{imu}{IMU}{Inertial Measurement Unit}
\newacronym{v2v}{V2V}{Vehicle-to-Vehicle}
\newacronym{v2i}{V2I}{Vehicle-to-Infrastructure}
\newacronym{i2i}{I2I}{Infrastructure-to-Infrastructure}
\newacronym{hdt}{HDT}{High-Definition Testfield}
\newacronym{HD maps}{HD maps}{High-Definition maps}
\newacronym{mAP}{mAP}{mean Average Precision}
\newacronym{sis}{SIS}{Separate Intersection Split}
\newacronym{eis}{EIS}{Equal Intersection Split}

\title{UrbanIng-V2X: A Large-Scale Multi-Vehicle, Multi-Infrastructure Dataset Across Multiple Intersections for Cooperative Perception}

\author{%
    \textbf{Karthikeyan Chandra Sekaran}$^{1}$\thanks{Equal contribution. Authors listed in alphabetical order.}\quad
    \textbf{Markus Geisler}$^{1}$\footnotemark[1]\quad
    \textbf{Dominik Rößle}$^{1}$\footnotemark[1] \\\\
    \textbf{Adithya Mohan}$^{1}$ \quad
    \textbf{Daniel Cremers}$^{2}$\quad
    \textbf{Wolfgang Utschick}$^{2}$\quad \\\\
    \textbf{Michael Botsch}$^{1}$\quad
    \textbf{Werner Huber}$^{1}$\quad
    \textbf{Torsten Schön}$^{1}$ \\\\
    \textbf{$^{1}$Technische Hochschule Ingolstadt} \quad
    \textbf{$^{2}$Technical University of Munich}
}

\begin{document}

\maketitle

\begin{abstract}\label{abstract}
    Recent cooperative perception datasets have played a crucial role in advancing smart mobility applications by enabling information exchange between intelligent agents, helping to overcome challenges such as occlusions and improving overall scene understanding. While some existing real-world datasets incorporate both vehicle-to-vehicle and vehicle-to-infrastructure interactions, they are typically limited to a single intersection or a single vehicle. A comprehensive perception dataset featuring multiple connected vehicles and infrastructure sensors across several intersections remains unavailable, limiting the benchmarking of algorithms in diverse traffic environments. Consequently, overfitting can occur, and models may demonstrate misleadingly high performance due to similar intersection layouts and traffic participant behavior. To address this gap, we introduce UrbanIng-V2X, the first large-scale, multi-modal dataset supporting cooperative perception involving vehicles and infrastructure sensors deployed across three urban intersections in Ingolstadt, Germany. UrbanIng-V2X consists of 34 temporally aligned and spatially calibrated sensor sequences, each lasting 20 seconds. All sequences contain recordings from one of three intersections, involving two vehicles and up to three infrastructure-mounted sensor poles operating in coordinated scenarios. In total, UrbanIng-V2X provides data from 12 vehicle-mounted RGB cameras, 2 vehicle LiDARs, 17 infrastructure thermal cameras, and 12 infrastructure LiDARs. All sequences are annotated at a frequency of 10 Hz with 3D bounding boxes spanning 13 object classes, resulting in approximately 712k annotated instances across the dataset. We provide comprehensive evaluations using state-of-the-art cooperative perception methods and publicly release the codebase, dataset, HD map, and a digital twin of the complete data collection environment via \href{https://github.com/thi-ad/UrbanIng-V2X}{\texttt{https://github.com/thi-ad/UrbanIng-V2X}}.
\end{abstract}

\begin{figure}[htbp]
    \centering
    \includegraphics[width=\textwidth]{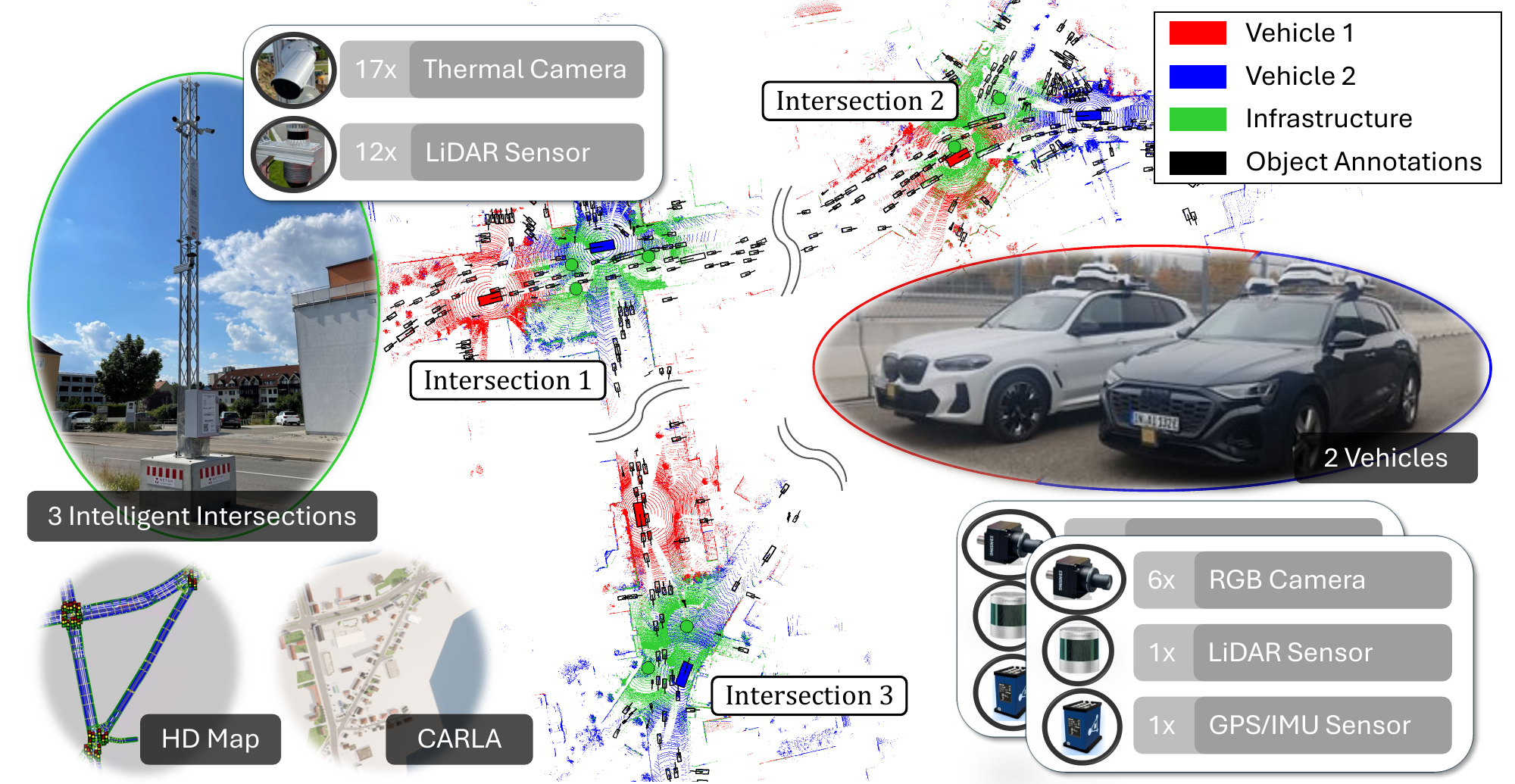}
    \caption{This illustration provides a comprehensive overview of the \textbf{UrbanIng-V2X} cooperative perception dataset environment. For each intersection, a globally fused point cloud of a representative scenario is visualized. Point clouds from individual agents are color-coded, highlighting two vehicles and sensor poles at three intersections as cooperation partners. Further, the complete sensor setup, along with a bird's-eye view of both the HD map and a high-fidelity CARLA map, is shown.}
    \label{fig:teaser}
\end{figure}

\section{Introduction}
\label{sec_introduction}
Reliable perception is fundamental to autonomous driving, particularly in complex urban intersections where a comprehensive understanding of the scene is essential for safe decision-making~\cite{selfdrivingcars-asurvey, StudySpeedPrediction, Tesla2016Crash}.
In such environments, single-agent systems are inherently constrained by their \gls{fov} and often fail to detect critical objects that are occluded by other vehicles and infrastructure~\cite{Gerner2024FCO}.
To address these limitations, recent research has increasingly focused on cooperative perception~\cite{Roessle2024TempEmbeddings, yang2023scope}, leveraging information sharing among multiple agents. 
To advance this paradigm, substantial effort has been invested in the development of real-world \gls{v2x} datasets~\cite{Yu2022DAIRV2X, Xiang2024V2XReal, zimmer2024tumtrafv2x}, despite significant challenges such as precise temporal and spatial synchronization, high financial and logistical costs, and the complexity of multi-agent hardware setups~\cite{Han2023CoopSurvey}.

However, none of the existing \gls{v2x} datasets capture urban scenarios involving multi-vehicle, multi-infrastructure setups across multiple intersections.
This combination is crucial to assess the scalability and real-world applicability of cooperative perception systems, especially in urban environments where heterogeneous sensor views and varying infrastructure layouts demand robust generalization and reliable performance.
To address this gap, we introduce the \textbf{UrbanIng-V2X} dataset---a large-scale cooperative perception dataset collected at three distinct urban intersections within the High-Definition Testfield~\cite{Botsch2023DataCollection, karthik2024}.
In this environment, we recorded 34 sequences, each lasting 20 seconds, involving two vehicles and up to three infrastructure-mounted sensor poles cooperating in coordinated scenarios.

\textbf{The main contributions of UrbanIng-V2X are as follows:}
\begin{itemize}
    \item UrbanIng-V2X is the first real-world cooperative perception dataset featuring multiple vehicles and extensive infrastructure sensing at three distinct urban intersections, see Figure~\ref{fig:teaser}. 
    
    \item UrbanIng-V2X introduces the largest number of cooperating sensors in real-world datasets to date and improves multi-modality by including thermal cameras. Each scenario involves 2 connected vehicles, each with 6 RGB cameras and a rooftop LiDAR, cooperating with up to 6 thermal cameras and 4 LiDARs mounted on 3 infrastructure poles per intersection. All sensors are spatially and temporally aligned, with additional sweep data provided for intermediate frames.
    
    \item The dataset supports a broad range of cooperative perception benchmark tasks, including 3D object detection, tracking, trajectory prediction, and localization. All sequences are annotated at 10~Hz with 3D bounding boxes for 13 object classes, totaling approximately 712k annotated instances. 

    \item A comprehensive benchmark evaluation is conducted using state-of-the-art (SOTA) algorithms for cooperative perception. The results highlight key challenges and opportunities across diverse sensor setups and intersection configurations.
    
    \item A developer toolkit is provided, including format converters for OpenCOOD~\cite{Xu2022OPV2V} and nuScenes~\cite{nuscenes}, for enabling integration with single- and multi-modal perception pipelines.
        
    \item The codebase and dataset, including \gls{HD maps} in Lanelet2 format~\cite{poggenhans2018lanelet2} and a geo-referenced CARLA~\cite{Dosovitskiy2017Carla} digital twin to support situation interpretation, synthetic data generation, and domain adaptation research, are publicly available.
    
\end{itemize}
\section{Related Work}
\label{sec_related_works}
\textbf{Single Agent Perception Datasets:} Large-scale single-vehicle datasets such as KITTI~\cite{Geiger2013IJRRKITTI}, nuScenes~\cite{nuscenes}, and the Waymo Open Dataset~\cite{Sun_2020_CVPR_Waymo} have been fundamental in advancing perception tasks.
They provide multi-modal sensor data captured from individual vehicles in diverse urban and suburban settings.
In contrast, datasets such as LUMPI~\cite{lumpi} and the TUMTraf Intersection Dataset~\cite{tumtraf} offer multi-modal sensor data, collected from infrastructure-mounted sensors.
While all these datasets support tasks such as 3D object detection and tracking, they inherently lack multi-agent interactions, limiting their utility in cooperative perception environments.

\textbf{Cooperative Perception Datasets: } Several synthetic datasets such as OPV2V~\cite{Xu2022OPV2V}, V2XSet~\cite{v2xvit}, and V2X-Sim~\cite{li2022v2xsim} have been developed in the past to explore cooperative perception with simulated multi-vehicle and infrastructure scenarios.
Although these datasets offer flexible and scalable environments, they fail to capture the full complexity and noise characteristics of real-world settings.
Consequently, several real-world datasets have emerged to support cooperative perception research. 
\textbf{V2V4Real}~\cite{Xu2023V2V4Real} enables \gls{v2v} cooperative perception across diverse driving scenes with rich annotations such as track IDs, but lacks infrastructure sensor data, limiting it to non-\gls{v2x} scenarios.
\textbf{DAIR-V2X-C}~\cite{Yu2022DAIRV2X} includes both \gls{v2i} data in 28 intersections---the most among existing datasets---but involves only one vehicle and lacks track IDs, HD maps, diverse sensors, and dense urban scenes. While its extension \textbf{V2X-Seq}~\cite{YuV2XSeq} adds track IDs and \gls{HD maps} for some sequences, the regional availability restrictions of DAIR-V2X-C~\cite{Yu2022DAIRV2X} and V2X-Seq~\cite{YuV2XSeq} limit international usability.
\textbf{TUMTraf-V2X}~\cite{zimmer2024tumtrafv2x} offers labeled single vehicle and infrastructure data at a single intersection with HD maps and day/night coverage.
However, its small scale and limited geographic diversity restrict its applicability.
\textbf{V2X-Real}~\cite{Xiang2024V2XReal} combines data from two vehicles and infrastructure, but lacks coverage of multiple intersections and HD maps.

We introduce UrbanIng-V2X to fill the gap of missing real-world datasets that cover a combination of multi-agent coordination, multi-modal sensing (including thermal cameras), and multiple intersection layouts. 
A detailed comparison to the existing datasets is shown in Table~\ref{tab:v2x_datasets}.

\begin{table}
\caption{Comparison of real-world cooperative \gls{v2x} datasets with the proposed UrbanIng-V2X dataset (I=Infrastructure, V=Vehicle). \dag Images are not published yet.}
\label{tab:v2x_datasets}
\centering
\begin{tabular}{p{0.146\textwidth}|
                >{\centering\arraybackslash}p{0.108\textwidth}|
                >{\centering\arraybackslash}p{0.118\textwidth}|
                >{\centering\arraybackslash}p{0.086\textwidth}|
                >{\centering\arraybackslash}p{0.107\textwidth}|
                >{\centering\arraybackslash}p{0.095\textwidth}||
                >{\centering\arraybackslash}p{0.116\textwidth}}
\toprule
\textbf{Property}      & \textbf{V2V4Real \cite{Xu2023V2V4Real}} & \textbf{DAIR-V2X-C\cite{Yu2022DAIRV2X}} & \textbf{V2X-Seq\cite{YuV2XSeq}} & \textbf{TUMTraf-V2X\cite{zimmer2024tumtrafv2x}} & \textbf{V2XReal \cite{Xiang2024V2XReal}} & \textbf{UrbanIng-V2X (ours)} \\
\midrule
Year                   &         2022         &     2022            &     2023            &      2024            &        2024       &       2025             \\
V2X                    &         V2V          &     V2I       &     V2I       &      V2I       &        V2V\&I  &       V2V\&I             \\
Intersections          &           \cellcolor{lightred}0          &     \cellcolor{lightgreen}28              &     \cellcolor{lightgreen}28              &        \cellcolor{lightred}1             &        \cellcolor{lightred}1          &        \cellcolor{lightgreen}3               \\
Vehicles               &           \cellcolor{lightgreen}2          &     \cellcolor{lightred}1               &     \cellcolor{lightred}1               &        \cellcolor{lightred}1             &        \cellcolor{lightgreen}2          &        \cellcolor{lightgreen}2               \\
RGB Images             &  \cellcolor{lightgreen}40k\dag &     \cellcolor{lightgreen}39k             &     \cellcolor{lightorange}15k             &         \cellcolor{lightred}5k           &        \cellcolor{lightgreen}171k        &      \cellcolor{lightgreen}81.6k               \\
IR Images              &          \cellcolor{lightred}0           &      \cellcolor{lightred}0              &      \cellcolor{lightred}0              &         \cellcolor{lightred}0            &        \cellcolor{lightred}0           &      \cellcolor{lightgreen}38.8k               \\
LiDAR frames           &     \cellcolor{lightorange}20k              &       \cellcolor{lightgreen}39k           &       \cellcolor{lightorange}15k           &        \cellcolor{lightred}2k            &       \cellcolor{lightgreen}33k          &      \cellcolor{lightgreen}27.2k               \\
3D Boxes              &    \cellcolor{lightorange} 240k             &       \cellcolor{lightgreen}464k          &       \cellcolor{lightred}10.45k          &        \cellcolor{lightred}29k           &      \cellcolor{lightgreen}1.2M         &      \cellcolor{lightgreen}712k               \\
Classes                &     \cellcolor{lightorange}5                &       \cellcolor{lightgreen}10            &       \cellcolor{lightgreen}9            &        \cellcolor{lightgreen}8             &       \cellcolor{lightgreen}10           &       \cellcolor{lightgreen}13               \\
Digital Twin           &       \cellcolor{lightred}No             &         \cellcolor{lightred}No          &         \cellcolor{lightred}No          &        \cellcolor{lightred}No            &      \cellcolor{lightred}No            &       \cellcolor{lightgreen}Yes      \\
Av. worldwide       &       \cellcolor{lightgreen}Yes          &         \cellcolor{lightred}No          &         \cellcolor{lightred}No          &        \cellcolor{lightgreen}Yes           &       \cellcolor{lightgreen}Yes          &         \cellcolor{lightgreen}Yes                  \\
HD Maps                &       \cellcolor{lightgreen}Yes            &         \cellcolor{lightred}No          &         \cellcolor{lightgreen}Yes          &        \cellcolor{lightgreen}Yes           &  \cellcolor{lightred}No  &            \cellcolor{lightgreen}Yes           \\
Attributes     &         \cellcolor{lightred}No          &      \cellcolor{lightred}No             &      \cellcolor{lightred}No             &        \cellcolor{lightgreen}Yes           &        \cellcolor{lightred}No          &            \cellcolor{lightgreen}Yes            \\
Track IDs              &       \cellcolor{lightgreen}Yes            &      \cellcolor{lightred}No             &      \cellcolor{lightgreen}Yes             &        \cellcolor{lightgreen}Yes           &        \cellcolor{lightgreen}Yes       &        \cellcolor{lightgreen}Yes                   \\
Traffic light          &      \cellcolor{lightred}No             &        \cellcolor{lightred}No            &        \cellcolor{lightred}No            &         \cellcolor{lightred}No           &       \cellcolor{lightred}No            &          \cellcolor{lightgreen}Yes                 \\
Sensors (I | V)       &         \cellcolor{lightred}0 | 8             &          \cellcolor{lightred}2 | 3           &          \cellcolor{lightred}2 | 3           &      \cellcolor{lightorange}5 | 4         &      \cellcolor{lightgreen}8 | 12        &    \cellcolor{lightgreen}  10 | 16    \\
City               &    Ohio         &       Beijing   &       Beijing   &       Munich     &      N.A.          &       Ingolstadt   \\
Country               &    USA         &       China   &       China   &       Germany     &      N.A.          &       Germany   \\

\bottomrule
\end{tabular}
\end{table}

\section{Dataset}
\label{sec_dataset}
Two connected vehicles and three smart infrastructures are used for data collection. Each intersection has 2 to 3 sensor poles. 
Each of these two vehicles is equipped with a high-precision \gls{imu}, a 128-ray high-end LiDAR sensor, and six \gls{fhd} RGB cameras oriented in six directions, providing a full \SI{360}{\degree} \gls{fov}.
The vehicles also receive \gls{rtk} correction data, achieving localization accuracy up to \SI{1}{\centi\meter}.
At each intersection, 2 to 3 poles are installed, each equipped with 1 to 3 VGA thermal cameras.
Additionally, six of the seven poles are equipped with a LiDAR setup, comprising a 64-ray midrange LiDAR and a 32-ray short-range blind-spot LiDAR.
Detailed sensor descriptions and \gls{fov} coverage for infrastructure sensors are provided in the supplementary material.

\subsection{Sensor Synchronization}
\label{subsec_sensor_syncronization}
The UTC clock is employed as a unified time reference to synchronize both vehicle- and infrastructure-mounted sensors.
The \gls{imu} synchronizes to UTC via GPS signals and acts as the \gls{ptp}~\cite{IEEEStandardPTP} master within the vehicle.
The camera capture card operates as a \gls{ptp} slave, inheriting the UTC reference from the \gls{imu}, while the LiDAR system obtains UTC timestamps independently through a dedicated GPS mouse.
Beyond time synchronization, sensor data acquisition is precisely coordinated.
The LiDAR is phase-locked so that its zero-degree orientation consistently aligns with integer multiples of its rotation cycle, establishing a deterministic angular reference.
Camera triggering is hardware-based and event-driven: instead of simultaneous image capture, each camera is triggered exactly when the LiDAR beam passes through its \gls{fov}.
This setup minimizes intermodal latency and ensures high-precision spatio-temporal alignment between LiDAR and camera data.

Each intersection is equipped with UTC-synchronized \gls{ptp} and \gls{ntp} time servers.
Thermal cameras are synchronized via the \gls{ntp} service, while all LiDAR units receive UTC timestamps through dedicated GPS mouses.
Similar to the vehicles, infrastructure LiDARs are phase-locked to ensure that their rotational cycles start and end simultaneously across devices.
Unlike vehicle-mounted cameras, the infrastructure thermal cameras operate asynchronously in free-run mode and are not externally triggered.
Due to the heterogeneous placement and \gls{fov} of these sensors, synchronized hardware triggering would not yield optimal alignment across all LiDAR-camera pairs.
Instead, during post-processing, the thermal image closest in time to each annotated LiDAR scan is selected.
With thermal cameras operating at 30~\gls{fps} and LiDARs at 20~\gls{fps}, every second LiDAR frame aligns with every third thermal camera frame with a maximum possible temporal misalignment of \SI{16.6}{\milli\second}, which corresponds to half the cycle time of the thermal cameras.

\subsection{Calibration}
\label{subsec_calibration}

\textbf{Intrinsic calibration:}  
All thermal and RGB cameras are calibrated using a checkerboard pattern.
Side-mounted vehicle cameras use a fisheye projection model due to their wide \gls{fov}, while all other cameras use a standard pinhole model.

\begin{figure}
    \centering
    \includegraphics[width=\linewidth]{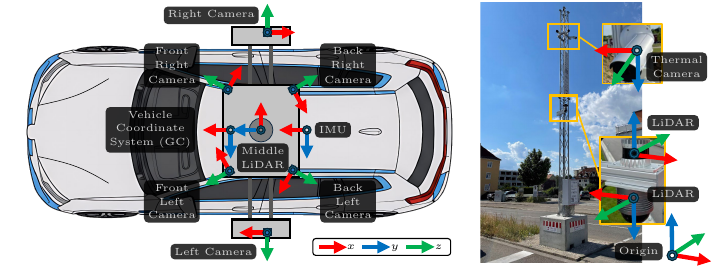}
    \caption{Sensor setup and coordinate frame. The left figure shows details of one vehicle, and the right figure shows details of one pole of a crossing. GC describes the geometric center.}
    \label{fig:coordinate_frame}
\end{figure}

\textbf{Extrinsic calibration:}  
Each sensor on the infrastructure (thermal camera, LiDAR) and vehicle (camera, LiDAR, \gls{imu}) has a local coordinate frame.
The vehicle coordinate frame is defined at the geometric center of the car.
Each intersection uses a fixed GPS location as its local origin, with the coordinate frame aligned to the East-North-Up (ENU) convention.
All coordinate systems use a right-handed convention.
Figure~\ref{fig:coordinate_frame} shows the coordinate systems for the vehicles and intersections.

The extrinsic transformation between the \gls{imu} and the vehicle coordinate frame ($\mathbf{T}_{\text{IMU} \rightarrow \text{Vehicle}}$) is precomputed using a total station. The \gls{imu} reports its pose in global coordinates via latitude, longitude, altitude, roll, pitch, and yaw, allowing precise computation of the vehicle’s global pose.
To estimate extrinsic transformation matrices between sensor pairs, we use a cone-shaped calibration target with a highly reflective marker in the center.
The cone is placed in multiple positions within the sensors' \gls{fov}.
Its global position is measured with a \SI{2}{\centi \meter} precise \gls{rtk} GPS device (Trimble SP80).
The reflective marker is manually annotated in both LiDAR point clouds and in camera images.
Using these annotations, we numerically optimize for transformation matrices by minimizing the reprojection error~\cite{nocedal2006numerical}.
This process yields the following extrinsic transformations:
\begin{itemize}
    \item $\mathbf{T}_{\text{Cam} \rightarrow \text{Global}}$ and $\mathbf{T}_{\text{LiDAR} \rightarrow \text{Global}}$ for infrastructure cameras and LiDARs respectively
    \item $\mathbf{T}_{\text{Cam} \rightarrow \text{Vehicle}}$ and $\mathbf{T}_{\text{LiDAR} \rightarrow \text{Vehicle}}$ for vehicle-mounted cameras and LiDARs respectively.
\end{itemize}

\begin{figure}
    \centering
    \includegraphics[width=\linewidth]{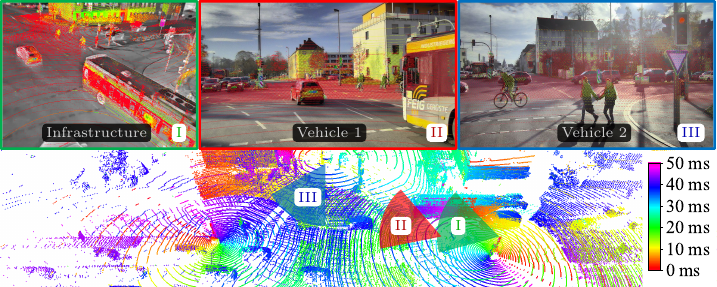}
    \caption{Result of the spatially calibrated and temporally aligned multi-modal sensor sources. The point cloud image highlights the time deviation in a globally fused cooperative LiDAR frame, particularly critical when LiDARs of multiple agents are capturing the same object. The top row shows the overlaid projections of the point cloud into three exemplary sensor perspectives.}
    \label{fig:calibration}
\end{figure}

\subsection{LiDAR Motion Compensation and Data Fusion}
\label{subsec_lidar_compensation_and_data_fusion}
Due to the rotational nature of the LiDAR sensor, each point within the point cloud is captured at a slightly different timestamp and vehicle pose. To accurately fuse point clouds from multiple sources, it is necessary to apply transformations on a per-point basis. 
Each point, captured at time $t$ with LiDAR pose $\mathbf{P}_t$, is transformed into the reference frame $\mathbf{P}_0$ at scan start time $t=0$ using the relative vehicle motion.
Such intra-scan motion compensation is crucial for achieving accurate fusion, especially in dynamic scenes where overlapping observations are acquired by multiple sensors undergoing relative motion.

Using the estimated intrinsic and extrinsic parameters, all vehicle and infrastructure LiDAR point clouds are transformed into a shared global coordinate frame.
This unified representation enables the projection of any LiDAR point into any camera image. Figure~\ref{fig:calibration} presents a fused multi-modal visualization of sensor data of a specific frame.
The bottom image displays the aggregated point cloud data from all vehicle-mounted and infrastructure-based LiDAR sensors.
Each point is color-coded based on its temporal offset from the start of the frame, highlighting that objects are observed by different LiDARs at varying timestamps.
Assuming a maximum object speed of \SI{50}{\kilo \meter \per \hour} and accounting for the worst-case sensor misalignment, the maximum spatial error within a frame is estimated to be \SI{0.7}{\meter}.
This error is unavoidable, as any object within the scene may move in arbitrary directions.
Additionally, Figure~\ref{fig:calibration} shows images from infrastructure and vehicle cameras with overlaid projections of the point cloud.

\subsection{Scenario Selection and Annotation}
\label{subsec_scenario_selection_and_annotation}

\begin{figure}[htbp]
     \centering
     \includegraphics[width=\linewidth]{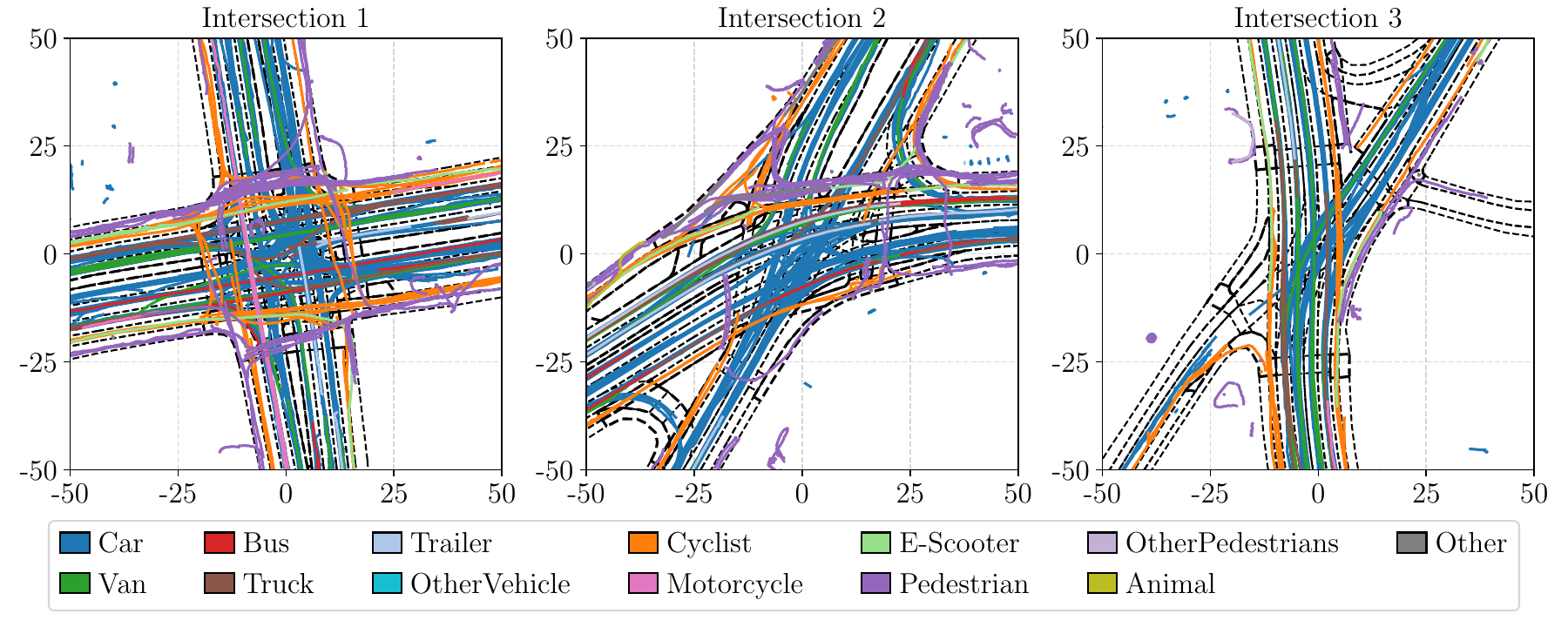}
     \caption{Trajectories projected onto the HD map of each intersection, color-coded by object category, illustrating the quality, density, and variation across the intersection layouts. In total, $2156$ trajectories of Intersection 1, $1895$ trajectories of Intersection 2, and $835$ tracks of Intersection 3 are shown.}
     \label{fig:trajectories on HDmap}
 \end{figure}

\begin{figure}[htbp]
    \centering
    \includegraphics[width=\linewidth]{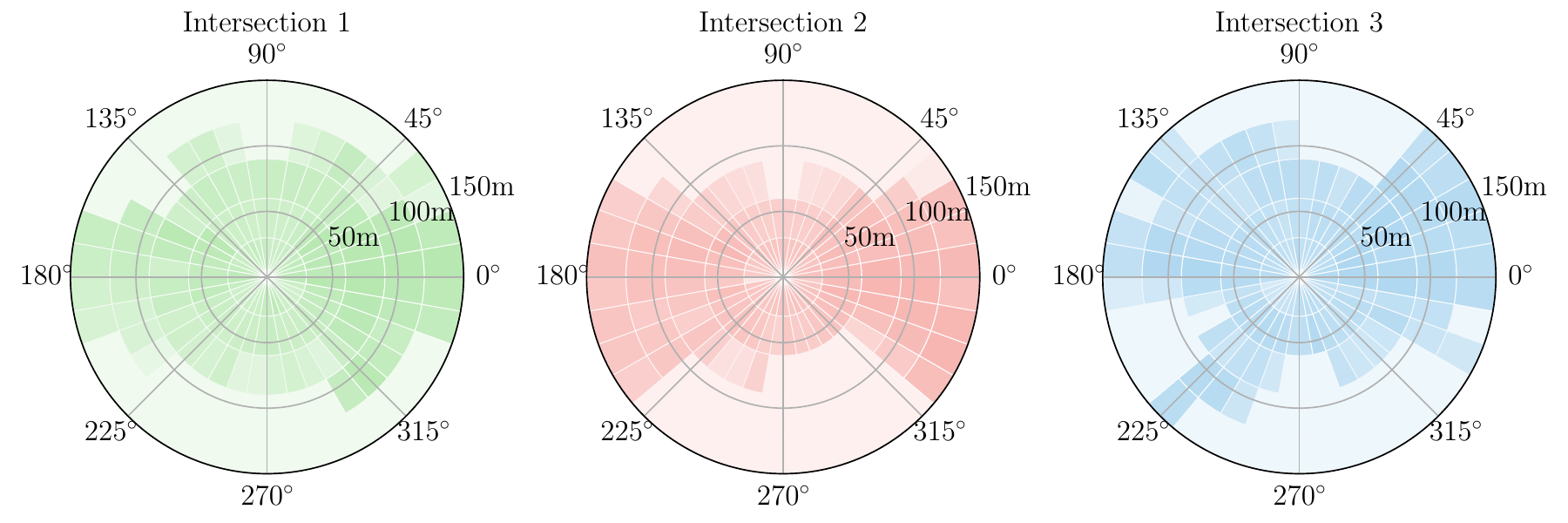}
    \caption{Polar density map showing object distributions by range and angle relative to vehicle agents, separated by intersection. Bin shading indicates object density, with $0^\circ$ aligned to the vehicle’s forward direction. Objects are densely distributed up to 150 meters. While high density along the vehicle axis is expected, the maps also reveal increased angular spread influenced by intersection layouts, which supports benchmark evaluations on generalization.}
    \label{fig:radial_plots}
\end{figure}

\begin{figure}[htbp]
    \centering
    \includegraphics[width=\linewidth]{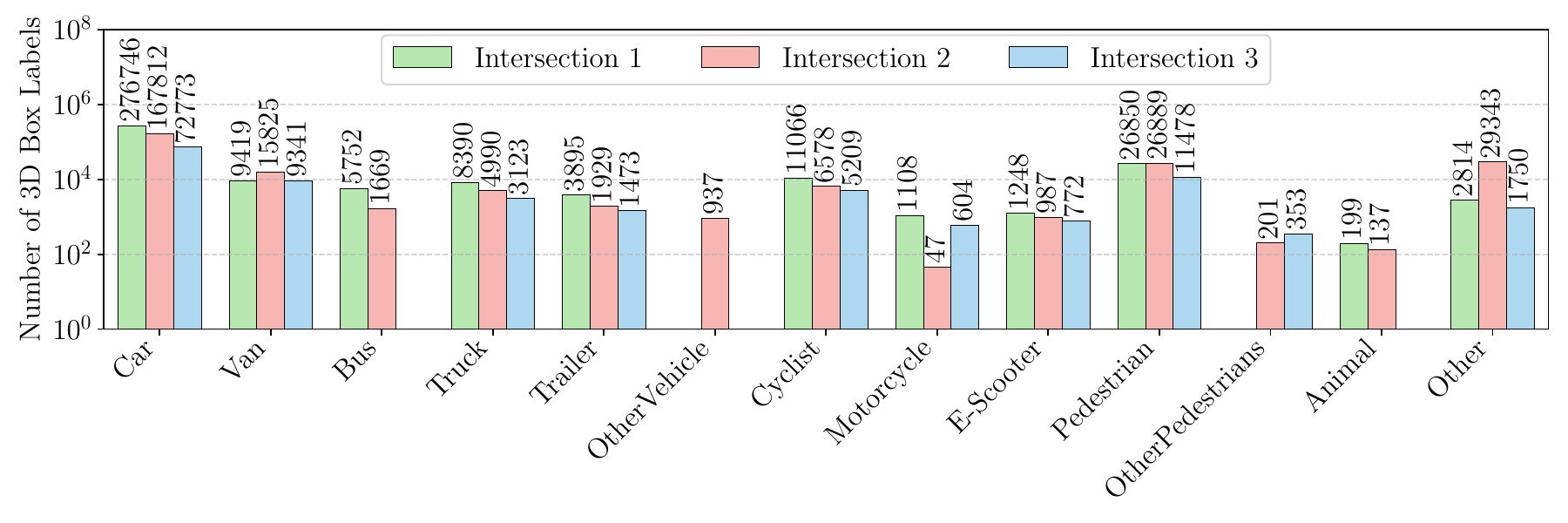}
    \caption{The number of annotated 3D bounding boxes for all 13 object classes, grouped by intersection. While cars constitute the highest amount of 3D annotations, also other categories such as pedestrians, and cyclists are well represented. The distribution is relatively evenly across the intersections. The only exception is OtherVehicle, which is predominantly represented by an excavator located exclusively in Intersection 2 and does not appear in other intersections.}
    \label{fig:object cagetegory distribution}
\end{figure}

\begin{figure}[htbp]
    \centering
    \begin{subfigure}[t]{0.32\linewidth}
        \centering
        \includegraphics[width=\linewidth]{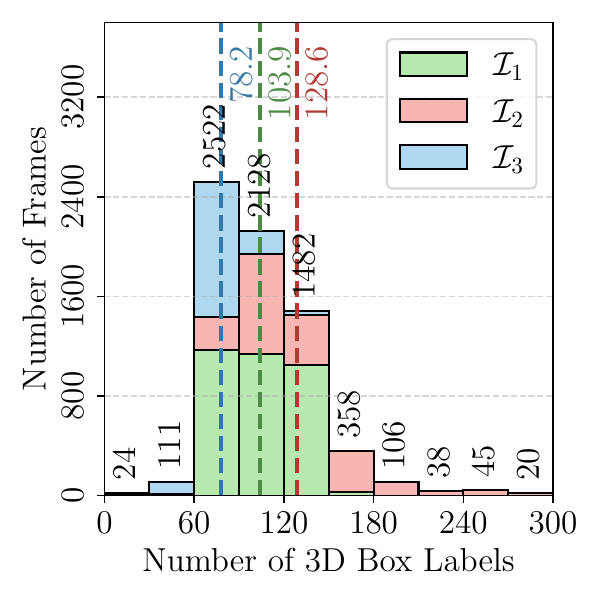}
        \caption{3D box labels per frame.}
        \label{fig:annotations_frames}
    \end{subfigure}
    \begin{subfigure}[t]{0.32\linewidth}
        \centering
        \includegraphics[width=\linewidth]{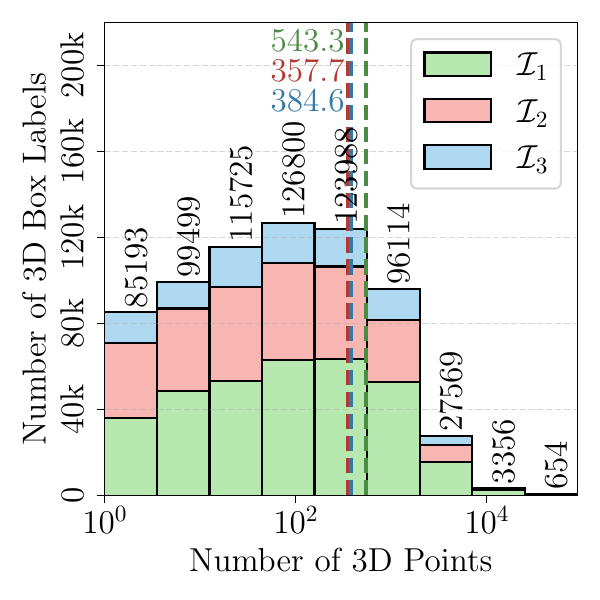}
        \caption{3D points per 3D box label.}
        \label{fig:annotations_label_points}
    \end{subfigure}
    \begin{subfigure}[t]{0.32\linewidth}
        \centering
        \includegraphics[width=\linewidth]{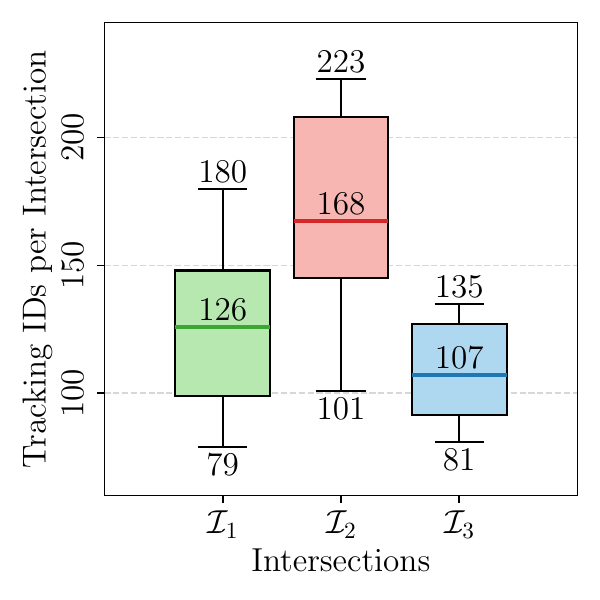}
        \caption{Tracking IDs per intersection.}
        \label{fig:tracking_ids}
    \end{subfigure}
    \caption{Intersection 1, 2, and 3 are abbreviated as $\mathcal{I}_1$, $\mathcal{I}_2$, $\mathcal{I}_3$, respectively. (a) Frames contain an average of $103.9$ objects in $\mathcal{I}_1$, $128.6$ in $\mathcal{I}_2$, and $78.2$ in $\mathcal{I}_3$. (b) 3D box labels contain an average of $543.3$ points in $\mathcal{I}_1$, $357.7$ in $\mathcal{I}_2$, and $384.6$ in $\mathcal{I}_3$, based on the fused point cloud. (c) Sequences contain a median of 126 tracks in $\mathcal{I}_1$, 168 in $\mathcal{I}_2$, and 107 in $\mathcal{I}_3$. Scene complexity, object density, and observation quality differ significantly across intersection types: $\mathcal{I}_1$ yields the highest point-level visibility per object, $\mathcal{I}_2$ features the most dynamic and densely populated scenarios, and $\mathcal{I}_3$ corresponds to sparser environments with reduced perceptual coverage.}
    \label{fig:Object and track number level statistics}
\end{figure}

The UrbanIng-V2X dataset is carefully curated from approximately eight hours of recorded data collected across three intersections.
Based on the raw recordings, we selected a set of 34 representative 20-second scenarios that capture diverse traffic situations and flow patterns, with a focus on varied vehicle behaviors and object categories.
The dataset comprises a wide range of illumination conditions, including 10 daytime, 5 cloudy, 6 moderate-light, 5 late-evening, and 8 nighttime scenarios.
All faces and license plates are anonymized using a Gaussian blur to comply with data protection regulations~\cite{gdpr2016}.
Annotations are applied to the fused point cloud data from all infrastructure sensors and vehicles, ensuring both spatial and temporal consistency across all modalities.
Data quality was rigorously validated through multiple rounds of quality control.
Each object is annotated with detailed 3D bounding boxes at a frequency of 10~Hz, specifying their spatial position $(x, y, z)$ and orientation.
Additionally, each object is assigned a unique tracking ID per sequence and categorized into one of 13 object classes. These annotations are further enriched with six attribute types, described in the supplementary materials.
Figure~\ref{fig:c2_multisensor_with_anns} illustrates one scenario sample from different sensor perspectives.
\begin{figure}
    \centering
    \includegraphics[width=\textwidth]{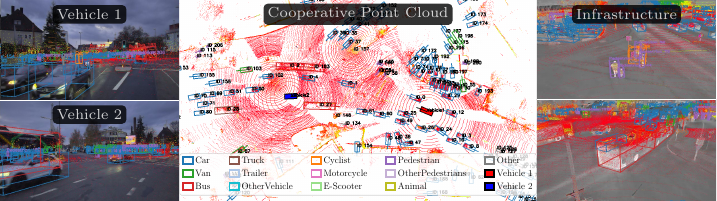}
    \caption{Projection of 3D annotations at one timestamp into three exemplary views: front left camera (left), bird’s-eye view fused point cloud (center), and two infrastructure cameras (right) are shown.} \label{fig:c2_multisensor_with_anns}
\end{figure}
Annotation characteristics are analyzed in detail across trajectory, frame, and object levels.
This includes visualizations such as trajectory overlays on HD maps, polar density maps, object category distributions, and statistics on object and track counts (see Figures~\ref{fig:trajectories on HDmap}, \ref{fig:radial_plots}, \ref{fig:object cagetegory distribution}, and \ref{fig:Object and track number level statistics}). 

\section{Tasks}
\label{sec_tasks}
UrbanIng-V2X provides comprehensive 3D annotations supporting multiple tasks, including object detection---the primary focus of this work---as well as object tracking, trajectory prediction, and localization.
The dataset further enables the evaluation of vehicle and infrastructure agents operating in various cooperative modes, allowing for evaluations of the performance of \gls{v2v}, \gls{v2i}, and \gls{i2i} (at a sensor pole level) at all three intersections.

\textbf{3D Object Detection.} For a structured analysis, we group the 13 annotated object categories into four superclasses: \textit{Vehicle} (Car, Van), \textit{Two-Wheelers} (Cyclist, Motorcycle, E-Scooter), \textit{Heavy Vehicle} (Truck, Bus, Trailer, Other Vehicle), and \textit{Pedestrian} (Pedestrian, OtherPedestrian).
The classes \textit{Animal} and \textit{Other} are excluded due to their low sample counts and high intra-class variability.
Bounding boxes beyond $\pm100$ meters in the x-direction and $\pm40$ meters in the y-direction are excluded~\cite{Xiang2024V2XReal}.
Furthermore, only objects with at least five LiDAR points in the fused point cloud of the selected agents are considered during both training and evaluation.
During training, ego agents are selected randomly to enable viewpoint diversity and improved model generalization.
For evaluation, we select one ego agent as an autonomous vehicle and the rest as collaborators, similar to V2X-Real~\cite{Xiang2024V2XReal}.
We report detection performance using the \gls{mAP} metric, evaluated at low IoU thresholds of $0.3$ and $0.5$ similar to V2XReal~\cite{Xiang2024V2XReal} and KITTI-360~\cite{KITTI360}.

\section{Experiments}
\label{sec_experiments}
We present 3D object detection LiDAR-only benchmark results in four strategies: no fusion, early fusion, late fusion, and intermediate fusion.
We use F-Cooper~\cite{FCooper}, AttFuse~\cite{Xu2022OPV2V}, V2X-ViT~\cite{v2xvit}, Where2Comm~\cite{Where2comm2022}, and CoBEVT~\cite{cobevt} for intermediate fusion.
All models are implemented using the PointPillars backbone~\cite{Lang2019PointPillars}.

\subsection{Dataset splits}\label{subsec: datasplit}
\label{subsec_dataset_splits}
To reliably assess the generalization capabilities of benchmark algorithms, it is crucial to employ dataset splitting strategies that prevent data leakage and enable fair evaluation.
Common approaches include frame-wise~\cite{zimmer2024tumtrafv2x, Xiang2024V2XReal, truckscenes2024} and sequence-wise splits~\cite{nuscenes}.
\textbf{Frame-wise splitting} distributes individual frames across the training, validation, and test sets by optimizing for equal characteristics of the data across the subsets.
However, this approach is prone to data leakage due to strong temporal correlation among frames and could lead to undetected overfitting and misleadingly high performance scores.
\textbf{Sequence-wise splitting} groups temporally consecutive frames (i.\,e., driving sequences) into the same set, potentially avoiding data leakage. This method ensures a more realistic evaluation of generalization but may result in less balanced distributions of data characteristics across the splits.

To account for limitations on existing split strategies, we propose two approaches, namely \textbf{\gls{eis}} and \textbf{\gls{sis}}.
\gls{eis} utilizes a sequence-dependent approach to fairly assess performance within known intersections.
To account for the possibility of sequence selection bias while maintaining representativeness, we define three randomized splits with non-overlapping validation and test sets across the splits.
Each split consists of 21 training, 6 validation, and 7 test sequences, proportionally distributed across all three intersections.
\textbf{\gls{sis}} leverages the presence of three distinct intersections in UrbanIng-V2X to enable \textbf{intersection-wise splitting}.
This approach strengthens independence by ensuring that all data from a given intersection appears exclusively in either the training, validation, or test split, thereby promoting generalization to entirely unseen locations.
\gls{sis} follows a leave-one-out scheme across the three intersections, with four configurations: $\text{\gls{sis}}_{1/2\text{vs.}3}$, 
$\text{\gls{sis}}_{2/3\text{vs.}1}$, and
$\text{\gls{sis}}_{1/3\text{vs.}2}$,
where indices denote the intersections used for training and validation versus testing. 4 sequences of intersection 1, 3 of intersection 2, and 3 of intersection 3 are consistently sampled for the validation split, if the intersection is part of the training split.

\subsection{Benchmark results}
\label{subsec_benchmark_results}
We use the $\text{\gls{sis}}_{1/2\text{vs.}3}$ split to benchmark on all above-mentioned SOTA algorithms to evaluate their performance on an unseen intersection.
The results are presented in Table~\ref{tab:ap_iou_comparison}.
Intermediate fusion methods generally outperform no fusion, late fusion, and early fusion, although the latter yields competitive performance.
Late fusion exhibits the weakest cooperative performance, indicating significant challenges in the association of agent-specific object lists. 
In contrast, AttFuse~\cite{Xu2022OPV2V} achieves the best overall performance, surpassing other methods by at least 0.6 \gls{mAP}@0.5. 
A category-wise comparison reveals that Vehicles is the best-performing class. In contrast, Heavy Vehicles, Pedestrians, and Two-Wheelers present greater challenges. We attribute this to the fact that Pedestrians are the smallest objects, while the Two-Wheelers and Heavy Vehicle superclasses exhibit the highest intra-class dimension variance, with a minimum of three original annotation categories.

\begin{table}
    \caption{Evaluation of SOTA algorithms using AP metrics on the $\text{\gls{sis}}_{1/2\text{vs.}3}$ split.}
    \label{tab:ap_iou_comparison}
    \centering
    \begin{tabular}{l|cc|cc|cc|cc||cc}
        \toprule
        \textbf{Method} & 
        \multicolumn{2}{c|}{\textbf{$\text{AP}_{\textit{Veh}}$}} & 
        \multicolumn{2}{c|}{\textbf{$\text{AP}_{\textit{HVeh}}$}} & 
        \multicolumn{2}{c|}{\textbf{$\text{AP}_{\textit{Ped}}$}} & 
        \multicolumn{2}{c||}{\textbf{$\text{AP}_{\textit{TWheel}}$}} & 
        \multicolumn{2}{c}{\textbf{\gls{mAP}}} \\
        \multicolumn{1}{r|}{\textbf{IoU}} & 0.3 & 0.5 & 0.3 & 0.5 & 0.3 & 0.5 & 0.3 & 0.5 & 0.3 & 0.5 \\
        \midrule
        No Fusion & 49.1 & 40.9 & 19.2 & 17.6 & 2.0 & 0.7 & 18.0 & 13.8 & 22.1 & 18.3 \\
        Early Fusion & 46.1 & 41.1 & 26.8 & 24.8 & 6.0 & 3.5 & 24.1 & 21.6 & 25.8 & 22.8 \\
        Late Fusion & 28.7 & 24.6 & 9.8 & 6.9 & 1.9 & 0.8 & 16.7 & 12.1 & 14.3 & 11.1 \\
        \midrule
        F-Cooper~\cite{FCooper} & 52.6 & 46.7 & 33.1 & 24.0 & 4.6 & 3.1 & 25.1 & 23.2 & 28.9 & 24.2 \\
        AttFuse~\cite{Xu2022OPV2V} & 52.7 & 47.6 & 34.1 & 27.8 & 7.1 & 4.6 & 23.7 & 22.1 & 29.4 & 25.5 \\
        V2X-ViT~\cite{v2xvit}& 52.0 & 46.2 & 32.5 & 22.2 & 5.8 & 3.5 & 19.7 & 18.0 & 27.5 & 22.5 \\
        Where2Comm~\cite{Where2comm2022} & 50.4 & 45.8 & 28.4 & 25.3 & 5.1 & 3.1 & 23.2 & 20.9 & 26.7 & 23.8 \\
            CoBEVT~\cite{cobevt} & 53.2 & 46.0 & 33.8 & 29.6 & 5.7 & 3.3 & 22.5 & 20.5 & 28.8 & 24.9 \\
        \bottomrule
    \end{tabular}
\end{table}

Further, detailed evaluation on the remaining \gls{sis} and \gls{eis} splits for the most recently published method CoBEVT~\cite{cobevt} are shown in Table~\ref{tab:split_results}.
The performance on \gls{eis}\textsubscript{avg} is 38.2 \gls{mAP}@0.5, while the average \gls{sis} performance drops to 24.2 \gls{mAP}@0.5. This indicates generalization issues that need to be solved for future cooperative perception applications, an open challenge that UrbanIng-V2X aims to address.

\begin{table}
    \caption{Evaluation of all combinations of \gls{eis} and \gls{sis} splits based on CoBEVT~\cite{cobevt}. \gls{eis}\textsubscript{avg} represents the averaged score across the three different \gls{eis} splits.}
    \label{tab:split_results}
    \centering
    \begin{tabular}{p{2.4cm}|cc|cc|cc|cc||cc}
        \toprule
        \multicolumn{1}{l|}{\textbf{Data split}} & 
        \multicolumn{2}{c|}{\textbf{$\text{AP}_{\textit{Veh}}$}} & 
        \multicolumn{2}{c|}{\textbf{$\text{AP}_{\textit{HVeh}}$}} & 
        \multicolumn{2}{c|}{\textbf{$\text{AP}_{\textit{Ped}}$}} & 
        \multicolumn{2}{c||}{\textbf{$\text{AP}_{\textit{TWheel}}$}} & 
        \multicolumn{2}{c}{\textbf{\gls{mAP}}} \\
        \multicolumn{1}{r|}{\textbf{IoU}} & 0.3 & 0.5 & 0.3 & 0.5 & 0.3 & 0.5 & 0.3 & 0.5 & 0.3 & 0.5 \\
        \midrule
        $\text{\gls{eis}}\textsubscript{avg}$ & 74.6 & 68.7 & 44.7 & 37.3 & 21.8 & 13.1 & 38.7 & 33.0 & 45.0 & 38.2 \\
        $\text{\gls{sis}}_{1/2\text{vs.}3}$ & 53.2 & 46.0 & 33.8 & 29.6 & 5.7 & 3.3 & 22.6 & 20.5 & 28.8 & 24.6 \\
        $\text{\gls{sis}}_{1/3\text{vs.}2}$ & 45.1 & 40.2 & 14.9 & 11.3 & 10.2 & 6.0 & 22.0 & 18.7 & 23.0 & 19.1 \\
        $\text{\gls{sis}}_{2/3\text{vs.}1}$ & 64.8 & 59.1 & 41.5 & 31.1 & 10.8 & 7.4 & 22.6 & 18.2 & 34.9 & 28.9 \\
        \bottomrule
    \end{tabular}
\end{table}

\section{Conclusion}
\label{sec_conclusion}

We present UrbanIng-V2X, the first large-scale cooperative perception dataset that integrates multi-vehicle, multi-infrastructure, and multi-sensor modalities across multiple urban intersections. By expanding the diversity of sensor types—including up to 12 RGB cameras, 6 thermal cameras, and 6 LiDAR sensors per scene—UrbanIng-V2X enables research into robust multi-modal, multi-view fusion. The dataset is uniquely designed to evaluate generalization by including both familiar and previously unseen intersection layouts, addressing a critical limitation in existing benchmarks.
Our initial baseline experiments with SOTA LiDAR-only cooperative detection models reveal a clear gap in generalization performance: while better results are achieved on known intersections, there is a significant drop of 14.0 \gls{mAP}@0.5 when models are applied to unseen environments. These results highlight the pressing need for research into models that generalize reliably across varied urban scenes.

To support the community, we release the complete dataset alongside a development kit, \gls{HD maps}, and a geo-referenced digital twin in CARLA to facilitate research in perception, tracking, prediction, and simulation. Despite its contributions, UrbanIng-V2X has certain limitations. The dataset is restricted to three intersections within Ingolstadt, Germany, and broader generalization will require extending the benchmark to more diverse urban settings and adverse weather conditions.
We invite the research community to use UrbanIng-V2X as a robust foundation for advancing cooperative perception and want to encourage research into generalization, data-efficient learning, and synthetic-to-real transfer techniques.

\section*{Acknowledgments}
The dataset was collected within the High-Definition Testfield, which was constructed within the KIVI project (45KI05C031) funded by the \textit{Federal Ministry for Digital and Transport of Germany (BMDV)}. This work was supported by the Hightech Agenda Bavaria, the SiRaMiS project (H2-F1116.IN/48/2), and the Bavarian Academic Forum - BayWISS, all funded by the \textit{Bavarian State Ministry of Science and the Arts (StMWK)}.

\bibliographystyle{plainnat}
\bibliography{references}

\newpage
\appendix

\section{Sensor setup}
\begin{figure}
    \centering
    \includegraphics[width=\linewidth]{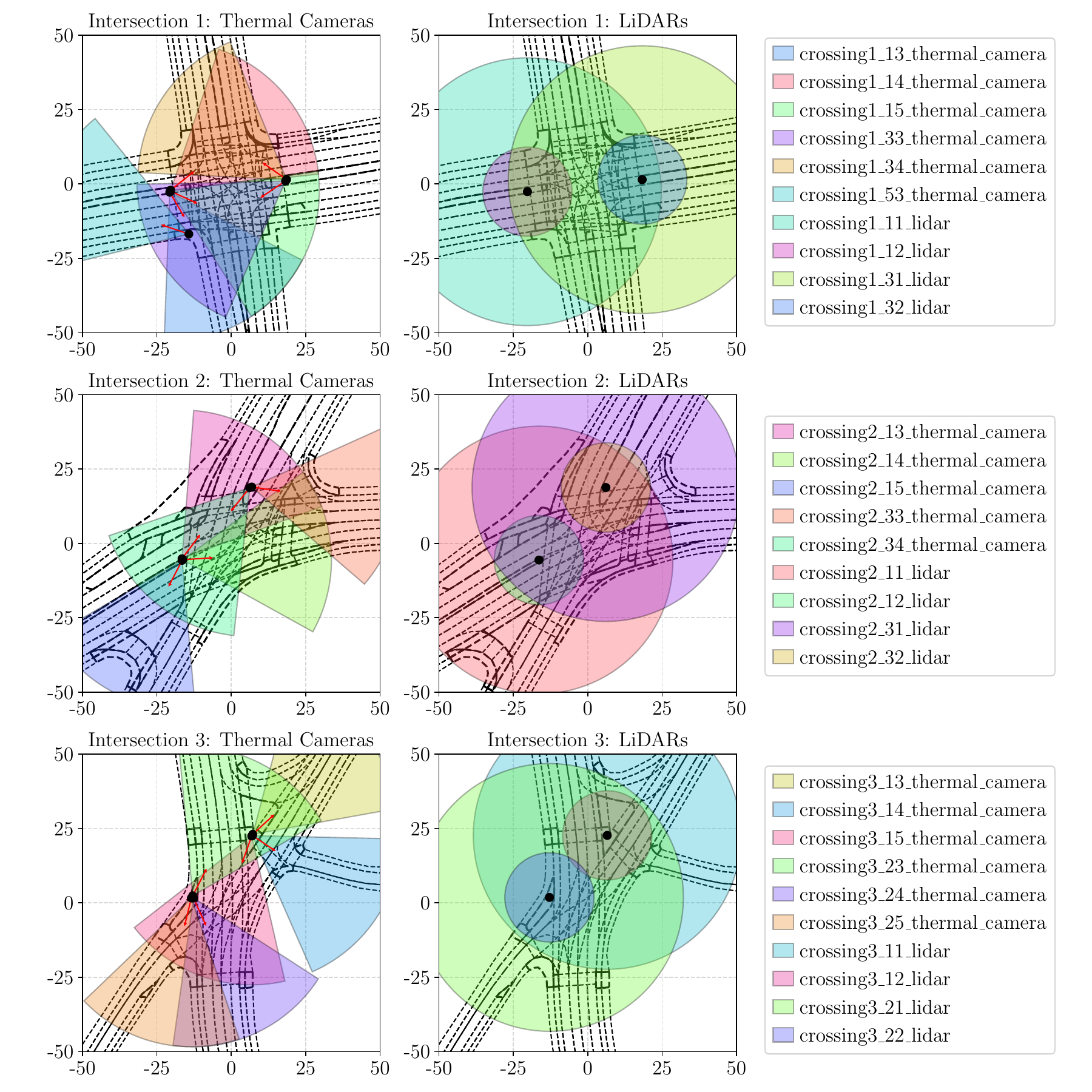}
    \caption{Sensor coverage for each intersection, with legend entries corresponding to folder names in the dataset.}
    \label{fig:sensors_fov_infrastructure}
\end{figure}

\begin{figure}
    \centering
    \includegraphics[width=.75\linewidth]{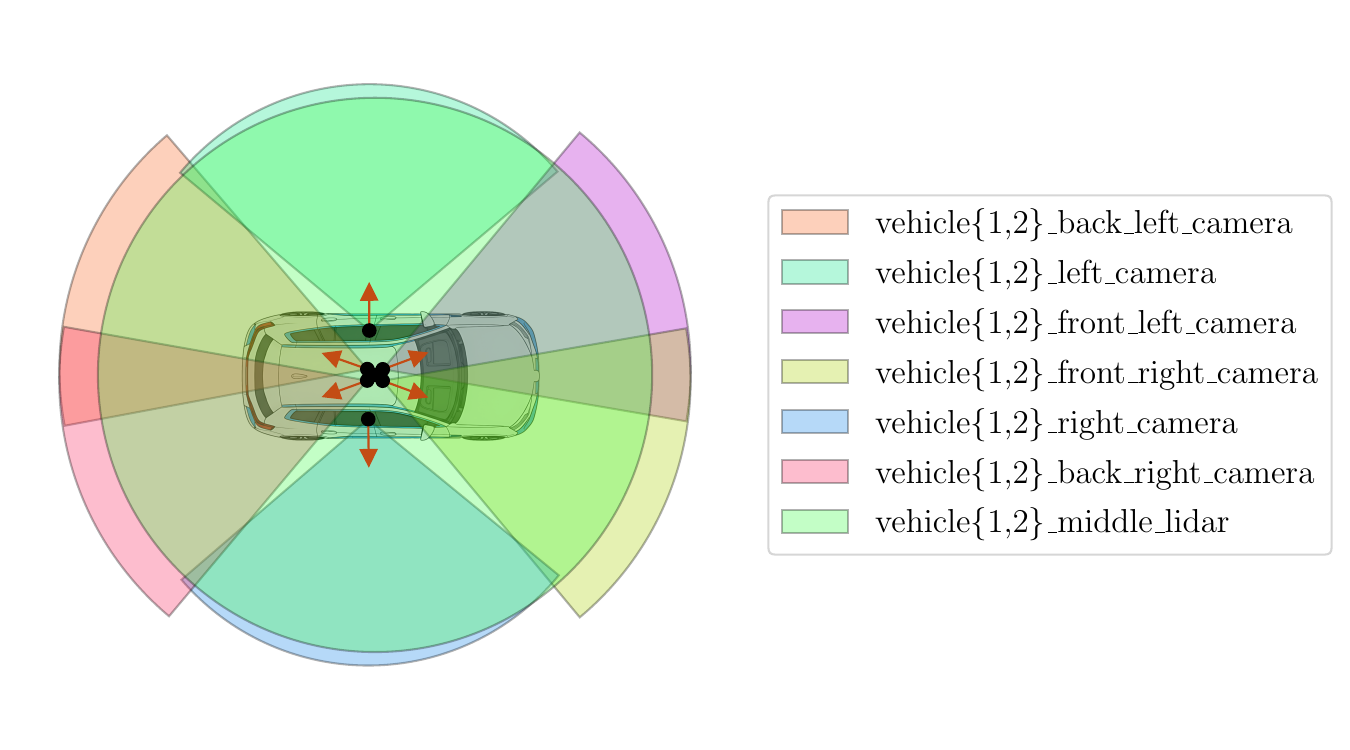}
    \caption{Sensor coverage for each vehicle, with legend entries corresponding to folder names in the dataset.}
    \label{fig:sensors_fov_vehicle}
\end{figure}

The UrbanIng-V2X dataset was collected using two vehicles and seven infrastructure-mounted sensor poles.
The setup spans three intersections: Intersection 1 is equipped with three sensor poles, while intersections 2 and 3 each have two sensor poles.
Intersection 1 includes 6 thermal cameras and 4 LiDAR sensors, Intersection 2 includes 5 thermal cameras and 4 LiDAR sensors, and Intersection 3 includes 6 thermal cameras and 4 LiDAR sensors.
Table~\ref{tab:sensor_setup_vehicle} and~\ref{tab:sensor_setup_infrastructure} provide an overview of the infrastructure and vehicle sensor specifications.
Figure~\ref{fig:sensors_fov_infrastructure} illustrates the \gls{fov} coverage of the thermal cameras and LiDAR sensors installed at each intersection.
Additionally, the \gls{fov} coverage of our vehicles for the RGB cameras and the LiDAR sensor is depicted in Figure~\ref{fig:sensors_fov_vehicle}.

\begin{table}[htb]
    \centering
    \caption{Vehicle sensor specifications (per vehicle)}
    \label{tab:sensor_setup_vehicle}
    \begin{tabular}{
               p{0.18\linewidth}|
               p{0.75\linewidth}}
        \toprule
        \textbf{Sensor} & \textbf{Details} \\
        \midrule
        RGB Cameras (6×) & Sensing GSML2 SG2-AR0233C-5200-G2A, 20 \gls{fps}, 1920 x 1080 resolution, \SI{60}{\degree} horizontal \gls{fov} (4x); \SI{100}{\degree} horizontal \gls{fov} (2x) \\
        LiDAR (1×) & Robosense Ruby Plus, 20 \gls{fps}, 128 rays, \SI{360}{degree} horizontal \gls{fov}, \SI{-25}{\degree} to \SI{15}{\degree} vertical \gls{fov}, $\leq$ \SI{240}{\meter} range at $\geq$ \SI{10}{\percent} reflectivity \\
        GPS/IMU (1×) & Genesys ADMA Pro+, 100 \gls{fps}, RTK correction, \SI{1}{\centi\meter} precise position data \\
        \bottomrule
    \end{tabular}
\end{table}

\begin{table}[htb]
    \centering
    \caption{Infrastructure sensor specifications (per intersection)}
    \label{tab:sensor_setup_infrastructure}
    \begin{tabular}{
               p{0.18\linewidth}|
               p{0.75\linewidth}}
        \toprule
        \textbf{Sensor} & \textbf{Details} \\
        \midrule
         Thermal Cameras (5–6×) & Axis Q1942-E, 30 \gls{fps}, 640 x 480 resolution, \SI{63}{\degree} horizontal \gls{fov} \\
        LiDARs (2×) & Ouster OS1-64 (Below Horizon) Rev 06, 20 \gls{fps}, 64~rays,  $\leq$~\SI{45}{\meter} range  at $\geq$~\SI{10}{\percent} reflectivity, \SI{360}{degree} horizontal \gls{fov}, \SI{-22.5}{\degree} to \SI{0}{\degree} vertical \gls{fov}\\ 
        LiDARs (2×) & Robosense Bpearl, 20 \gls{fps}, 32 rays,  $\leq$~\SI{30}{\meter} range  at $\geq$~\SI{10}{\percent} reflectivity blind spot sensor, \SI{360}{degree} horizontal \gls{fov}, \SI{-90}{\degree} to \SI{0}{\degree} vertical \gls{fov} \\
        \bottomrule
    \end{tabular}
\end{table}

\section{Data annotation}

\subsection{Annotation process}
The annotations underwent a multi-stage quality assurance process. 
After the initial annotation phase, in total, three review cycles with a manual refinement of bounding boxes by a professional annotation company were performed. At each stage, independent reviewers reported errors to enhance the precision of bounding boxes, object trajectories, and orientation estimates across the dataset.

\subsection{Object classes and object attributes}\label{sec:obj_classes_and_attributes}
\begin{figure}[ht]
    \centering
    \includegraphics[width=\linewidth]{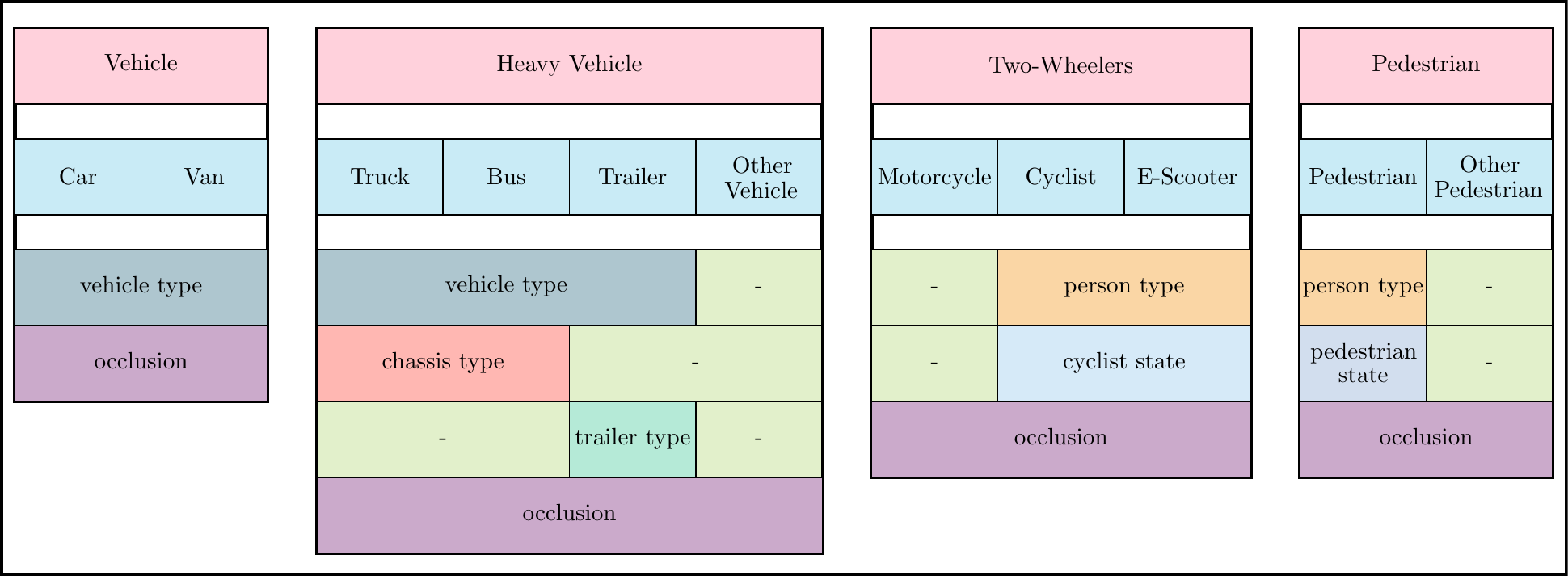}
    \caption{The first row displays the superclasses used for training the multi-object detectors. The second row shows the individual object classes grouped under their respective superclasses. The third row illustrates the attributes associated with each object class, represented by different color codes. The object classes Animal and Other are excluded from this overview, as they are not assigned to any superclass and are only annotated with the occlusion attribute.}
    \label{fig:class_attributes}
\end{figure}

\begin{figure}[ht]
    \centering
    \includegraphics[width=\linewidth]{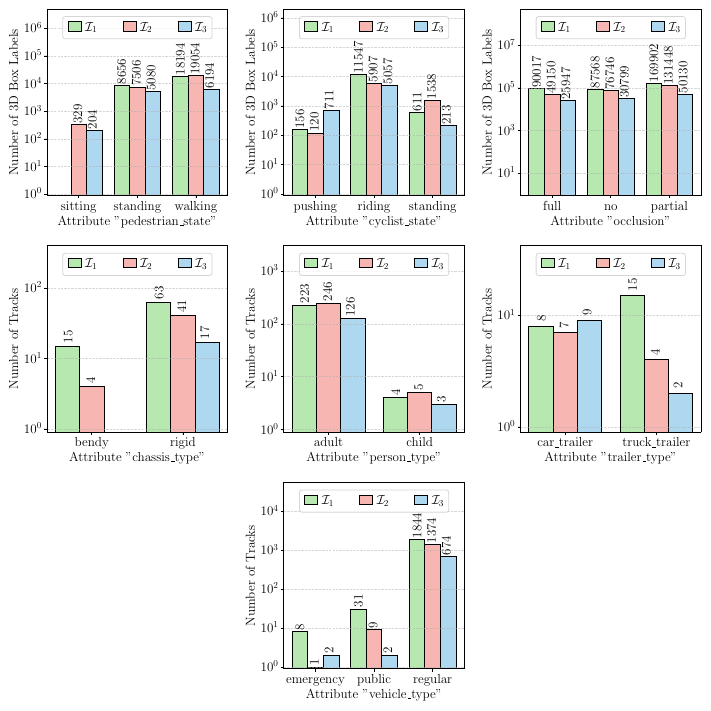}
    \caption{Visualization of all attributes and the frequency of their subcategories. We abbreviate Intersection 1, Intersection 2, and Intersection 3 as $\mathcal{I}_1$, $\mathcal{I}_2$, and $\mathcal{I}_3$, respectively. The attributes pedestrian\_state, cyclist\_state, and occlusion are frame-based. All other attribute types are track-based.}
    \label{fig:attributes_counts}
\end{figure}

In addition to class labels, we assigned semantic attributes to all annotated objects to capture more detailed characteristics and behavioral states.
For the purpose of benchmark evaluation, we grouped specific and closely related object classes into superclasses to perform a more structured detection task.
Figure~\ref{fig:class_attributes} illustrates the structure of these superclasses, their associated object types, and the attribute types applicable to each object type.
The object types Animal and Other are not included in the figure, as they were underrepresented in the dataset and thus not grouped into any superclasses.
However, both object types are also annotated with the occlusion attribute.
Figure~\ref{fig:attributes_counts} provides an overview of all attribute types and their respective subcategories, along with the frequency of their occurrences in the dataset.

\subsection{Data anonymization}\label{subsec:data_anonymization}
To ensure privacy compliance, we anonymized the RGB camera data of our dataset.
The dataset comprises a total of $163.200$ full HD RGB images, recorded at 20 \gls{fps} across all sequences.
All visible faces and license plates were annotated with 2D bounding boxes and anonymized using a Gaussian blur with a 75×75 kernel.
The annotations are publicly available in our Git repository to support advanced techniques such as inpainting or synthetic replacement.

\subsection{Extended annotation statistics}
\begin{figure}
    \centering
    \includegraphics[width=\linewidth]{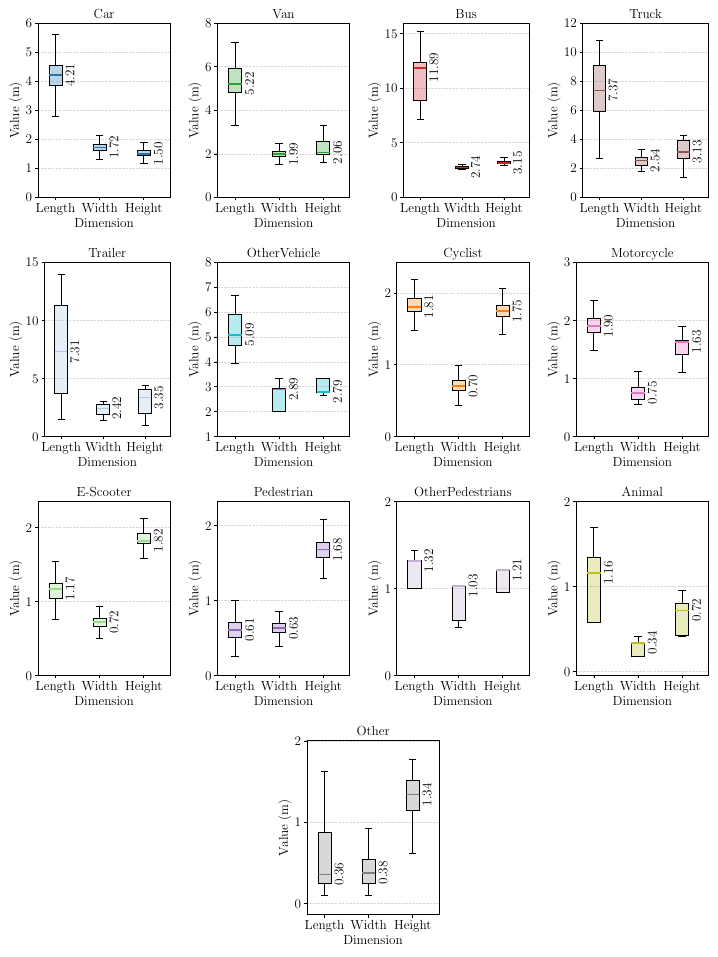}
    \caption{Distribution of object dimensions (length, width, height) for each dataset class. Each box plot summarizes the statistical spread of object dimensions per class, highlighting inter-class variation.}
    \label{fig:lwh_plot}
\end{figure}

\begin{figure}
    \centering
    \includegraphics[width=\linewidth]{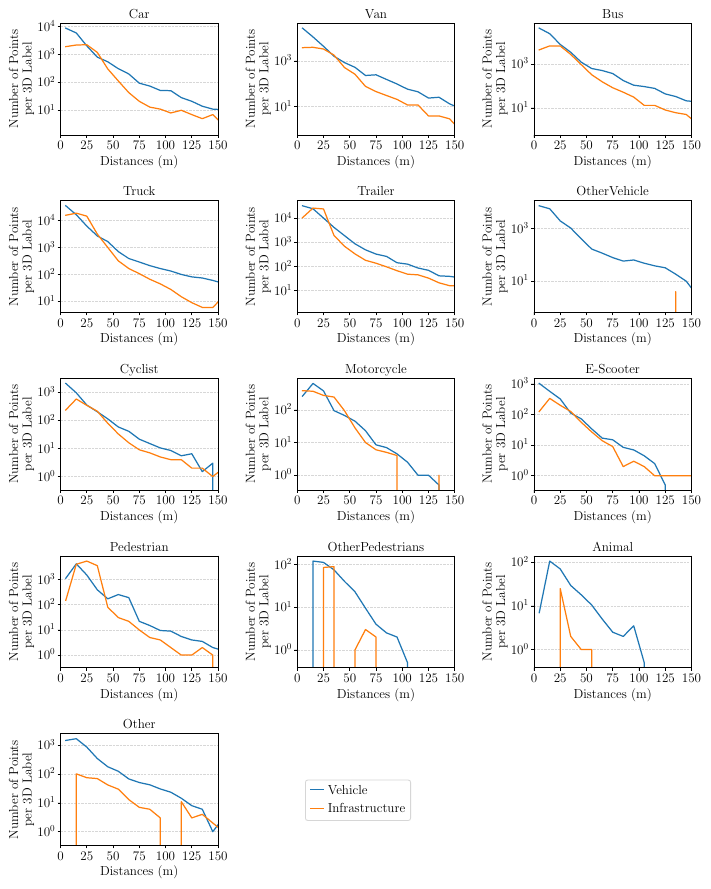}
    \caption{Comparison of the number of LiDAR points per 3D annotation across various distances. The infrastructure data is based on the fused point cloud in the local intersection origins. The vehicle values represent the average across both vehicle agents with respect to their vehicle coordinate frames. Specifically for the infrastructure plots, some classes exhibit irregular point density trends. The reasons are the static nature of the sensors and the sparse distribution of annotated instances across certain distance intervals.}
    \label{fig:points_per_distance}
\end{figure}

To provide further insights into the dataset, we present additional statistics for all annotated object classes.
Figure~\ref{fig:lwh_plot} shows the distributions of the object dimensions for all provided classes of our dataset. 
These distributions also reveal intra-superclass variations, for instance, highlighting the significant variance observed among classes within the Heavy Vehicle category.
Figure~\ref{fig:points_per_distance} presents the average number of LiDAR points captured per 3D bounding box across varying distances. Each subplot represents a specific object class and compares the point density of vehicle-mounted and infrastructure-mounted LiDAR sensors per agent.
While the superclasses Vehicle and Heavy Vehicle have densities up to 100 points per object at the benchmark range of \SI{100}{\meter}, the superclasses Pedestrian and Two-Wheelers show significant drops at ranges of approximately \SI{70}{\meter}.

\section{Experiments}
\subsection{Computer resources}
The experiments and computations described in this work were performed on a workstation equipped with an NVIDIA RTX A6000 GPU and an AMD Ryzen Threadripper PRO 5955WX processor with 16 cores, running Ubuntu 22.04.5 LTS.

\subsection{Implementation details}
\textbf{Multi-class detection.}
Our dataset includes 13 distinct object classes: Car, Van, Bus, Truck, Trailer, Vehicle, Cyclist, Motorcycle, E-Scooter, Pedestrian, OtherPedestrians, Animal, Other.
These classes were selected to reflect the diversity of road users in urban environments and to enable comprehensive multi-class detection.
We follow the standardized training procedure of the OpenCOOD~\cite{Xu2022OPV2V} framework and follow the approach of V2XReal~\cite{Xiang2024V2XReal}, leveraging the OpenPCDet framework~\cite{openpcdet2020} to enable multi-class evaluation.
All models are trained for 60 epochs with a batch size of 4.
We use the Adam optimizer with a learning rate of \(1 \times 10^{-3}\) and a weight decay of \(\sigma = 10^{-4}\).
A cosine annealing learning rate schedule is applied, starting with a warm-up phase over the first 10 epochs. During this phase, the learning rate increases from \(2 \times 10^{-4}\) to its peak and gradually decays to \(2 \times 10^{-5}\) by the final epoch.
Model configurations, hyperparameters, and training setups for all approaches are provided.
For evaluation, we selected the model checkpoint corresponding to the best validation performance, evaluated every 10 epochs.

\textbf{Model training time.}
Since our dataset enables focusing on different intersections across training and testing, we use the $\text{\gls{sis}}_{1/2\text{vs.}3}$ dataset split and the overall best-performing CoBEVT~\cite{cobevt} model for training time estimates.

The $\text{\gls{sis}}_{1/2\text{vs.}3}$ split includes:
\begin{itemize}
    \item Training set: 13 sequences from Intersection 1 and 7 sequences from Intersection 2 (total: 20 sequences).
    \item Validation set: 4 sequences from Intersection 1 and 3 sequences from Intersection 2 (total: 7 sequences).
    \item Test set: All 7 sequences from Intersection 3.
\end{itemize}
Training is parallelized using 16 PyTorch data workers (equal to the number of available CPU cores) for efficient data loading and augmentation.
As a representative example, training the CoBEVT model on the $\text{\gls{sis}}_{1/2\text{vs.}3}$ split requires approximately 18 hours to complete 60 epochs.

\subsection{Further benchmark results}
We further present the results of multiple benchmark methods for a single \gls{eis} split. All intermediate fusion models perform better within known intersections than on the $\text{\gls{sis}}_{1/2\text{vs.}3}$ split (Table~\ref{tab:ap_iou_comparison}). While AttFuse~\cite{Xu2022OPV2V} is the best-performing method when generalizing to an unknown intersection, V2X-ViT~\cite{v2xvit} attains the highest performance for the  \gls{eis} split evaluation in Table \ref{tab:further_ap_iou_comparison}.

\begin{table}[h]
    \caption{Evaluation of SOTA algorithms using AP metrics on a single random \gls{eis} split.}
    \label{tab:further_ap_iou_comparison}
    \centering
    \begin{tabular}{l|cc|cc|cc|cc||cc}
        \toprule
        \textbf{Method} & 
        \multicolumn{2}{c|}{\textbf{$\text{AP}_{\textit{Veh}}$}} & 
        \multicolumn{2}{c|}{\textbf{$\text{AP}_{\textit{HVeh}}$}} & 
        \multicolumn{2}{c|}{\textbf{$\text{AP}_{\textit{Ped}}$}} & 
        \multicolumn{2}{c||}{\textbf{$\text{AP}_{\textit{TWheel}}$}} & 
        \multicolumn{2}{c}{\textbf{\gls{mAP}}} \\
        \multicolumn{1}{r|}{\textbf{IoU}} & 0.3 & 0.5 & 0.3 & 0.5 & 0.3 & 0.5 & 0.3 & 0.5 & 0.3 & 0.5 \\
        \midrule
        F-Cooper~\cite{FCooper} & 70.4 & 62.5 & 42.6 & 32.2 & 11.5 & 5.1 & 23.1 & 20.1 & 36.9 & 30.0 \\
        AttFuse~\cite{Xu2022OPV2V} & 65.8 & 56.9 & 48.0 & 36.8 & 11.1 & 5.9 & 19.1 & 15.7 & 36.0 & 28.8 \\
        V2X-ViT~\cite{v2xvit} & 73.2 & 65.0 & 48.7 & 38.6 & 17.2 & 8.7 & 28.4 & 23.0 & 41.9 & 33.8 \\
        Where2Comm~\cite{Where2comm2022} & 67.9 & 59.5 & 40.0 & 31.4 & 12.3 & 6.3 & 17.8 & 14.8 & 34.5 & 28.0 \\
        CoBEVT~\cite{cobevt} & 71.8 & 63.4 & 46.6 & 35.6 & 18.5 & 9.4 & 29.5 & 22.5 & 41.7 & 33.2 \\
        \bottomrule
    \end{tabular}
\end{table}

\section{Limitations} 
The presented dataset comprises three intersections in Ingolstadt, Germany, offering a more diverse setting than existing real-world cooperative perception benchmarks with multiple vehicles and multiple infrastructure poles. 
Ingolstadt is one of the few cities in Germany with a permanent, multi-intersection V2X infrastructure deployment at this scale, making it uniquely suited for collecting a dataset of this complexity. The selected locations were deliberately chosen along major arterial roads that reflect common traffic dynamics, infrastructure layouts, and occlusion patterns typically observed in many European metropolitan areas.
While this contributes a step forward in promoting generalization challenges in SOTA cooperative perception algorithms, further extensions to intersections across a wider range of cities and urban topologies could support broader applicability and robustness. Future research could also focus on capturing data under adverse weather conditions such as rain, fog, or snow to improve environmental diversity. With a total sequence length of approximately 11 minutes for each agent, UrbanIng-V2X achieves a per-agent duration comparable to other V2X datasets listed in Table~\ref{tab:v2x_datasets}, though it remains smaller than SOTA single-agent autonomous driving datasets. Even though the data collection process and objectives of cooperative perception datasets differ fundamentally from single-agent recordings, a long-term goal for the field is to scale their raw recording durations toward the levels of SOTA single-agent autonomous driving datasets.
Annotations were performed using LiDAR data to ensure high-precision depth estimation.
Objects visible exclusively in camera sensors, for example, at great distances without corresponding LiDAR points, may not be annotated.
Despite extensive quality assurance, including rounds of manual annotation review, the procedure itself inherently carries a risk of human error.

\section{Societal impact}
Cooperative perception offers significant potential to improve situational awareness and safety in autonomous systems, particularly within complex urban environments.
By enabling vehicles to share sensor data and jointly interpret surroundings, it addresses key limitations of isolated single-agent autonomy.
However, this shift toward multi-agent cooperation introduces new challenges, including the need for a reliable and secure communication infrastructure ~\cite{CAV_Opportunities_and_Challenges}.
Beyond technical concerns, these systems may pose broader risks to personal privacy and the autonomy of individual drivers, as increased connectivity could enable persistent monitoring, centralized control, or unintended surveillance.
Moreover, while cooperative perception could yield substantial benefits in transportation safety, efficiency, and comfort, its societal value depends on the equitable deployment of autonomous technologies.
Without deliberate policy and investment, these technologies risk deepening existing disparities by primarily benefiting higher-income populations~\cite{SocietalBenefits}.
As autonomous vehicles are projected to account for up to 30 percent of urban traffic by 2030, technology~\cite{VisionDigitalizedAutomotiveIndustry2030}, with connectivity identified as a key enabler, it is critical that their development is guided by supportive regulatory frameworks that safeguard the broader public interest.

\section{Dataset visualization}\label{sec:dataset_visualiation}
This section provides a detailed overview of our dataset. The trajectories of all intersection sequences are visualized in Section~\ref{subsec:intersection1_trajectory_vis},~\ref{subsec:intersection2_trajectory_vis},~\ref{subsec:intersection3_trajectory_vis}.
In addition, for each intersection we show one representative frame from all sensor perspectives in Section~\ref{subsec:frame_vis_i1},~\ref{subsec:frame_vis_i2}, and~\ref{subsec:frame_vis_i3}.

\subsection{Intersection 1 trajectory visualization}\label{subsec:intersection1_trajectory_vis}
\begin{figure}[ht]
    \centering
    \includegraphics[width=\linewidth]{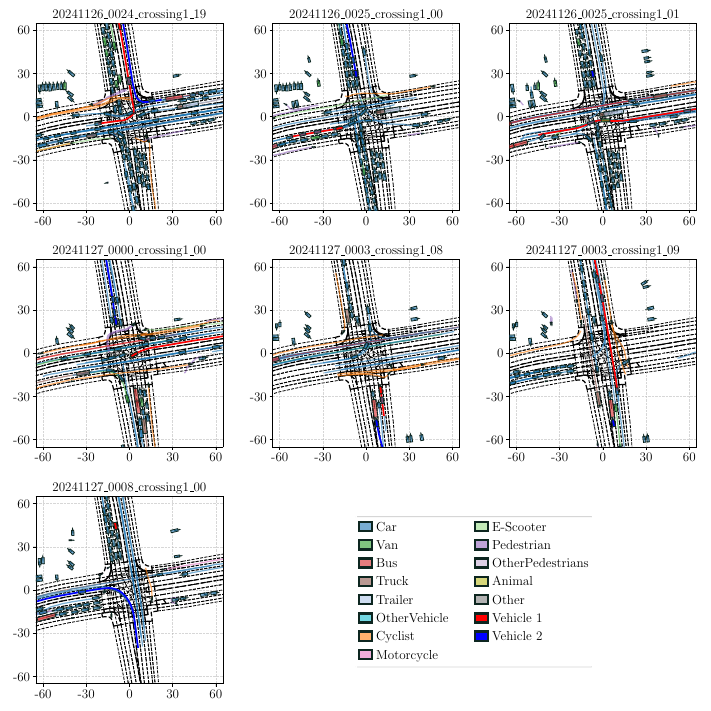}
    \caption{Visualization of trajectories at Intersection 1 across sequences 1-7. Each subplot shows the trajectories of all annotated object classes for a sequence. The sequence names correspond to the original folder names in the dataset.}
    \label{fig:merge_tracks_c1b}
\end{figure}

\begin{figure}[ht]
    \centering
    \includegraphics[width=\linewidth]{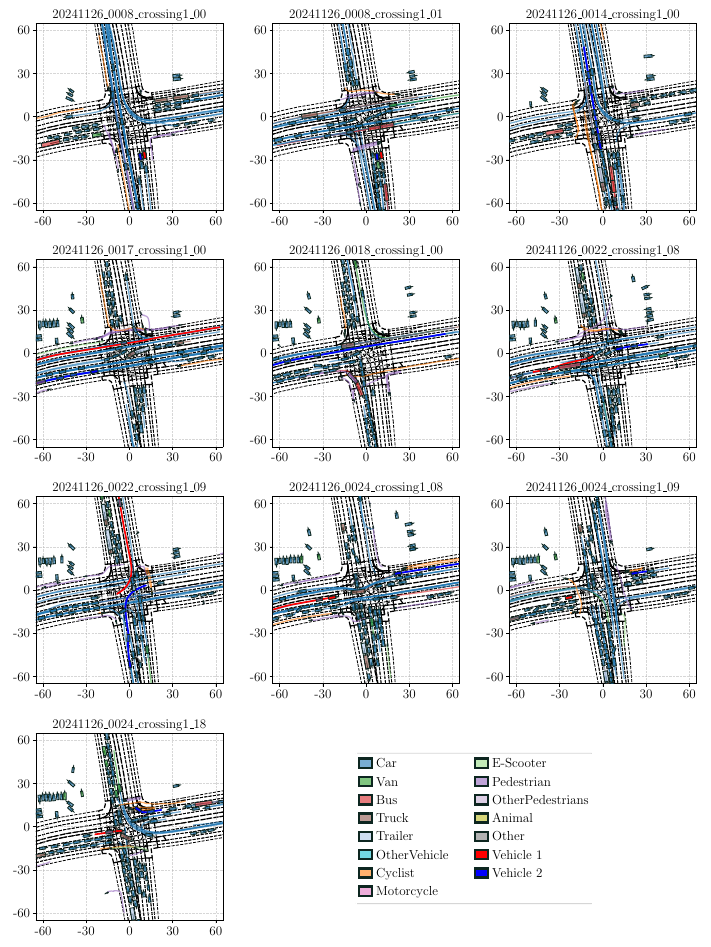}
    \caption{Visualization of trajectories at Intersection 1 across sequences 8-17. Each subplot shows the trajectories of all annotated object classes for a sequence. The sequence names correspond to the original filenames in the dataset.}
    \label{fig:merge_tracks_c1a}
\end{figure}

\clearpage
\newpage
\subsection{Intersection 2 trajectory visualization}\label{subsec:intersection2_trajectory_vis}
\begin{figure}[ht]
    \centering
    \includegraphics[width=\linewidth]{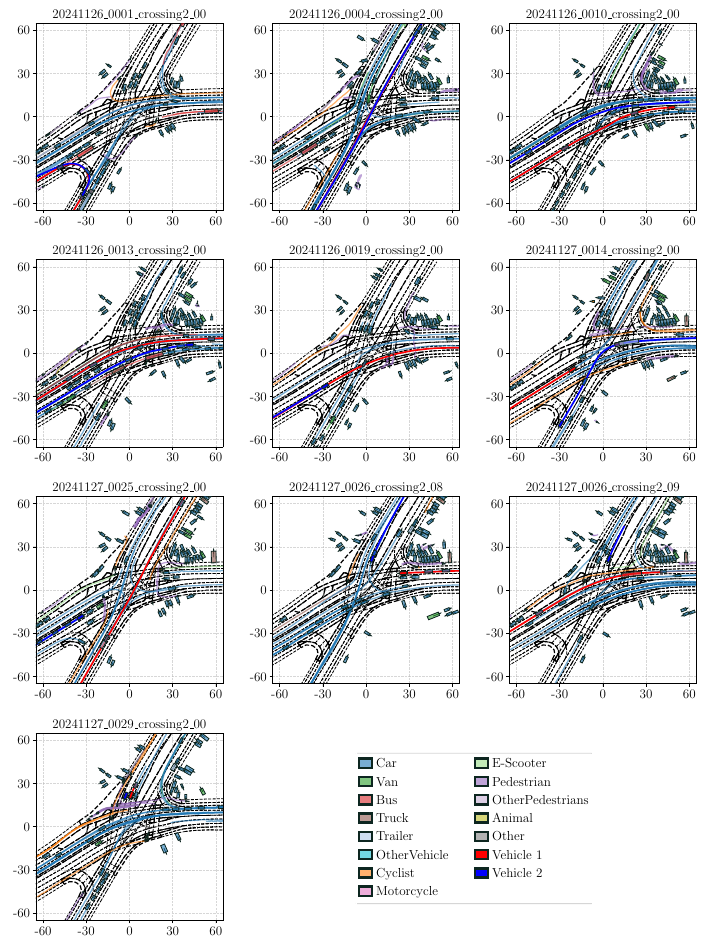}
    \caption{Visualization of trajectories at Intersection 2 across all sequences. Each subplot shows the trajectories of all annotated object classes for a sequence. The sequence names correspond to the original filenames in the dataset.}
    \label{fig:merge_tracks_c2}
\end{figure}

\clearpage
\newpage
\subsection{Intersection 3 trajectory visualization}\label{subsec:intersection3_trajectory_vis}
\begin{figure}[ht]
    \centering
    \includegraphics[width=\linewidth]{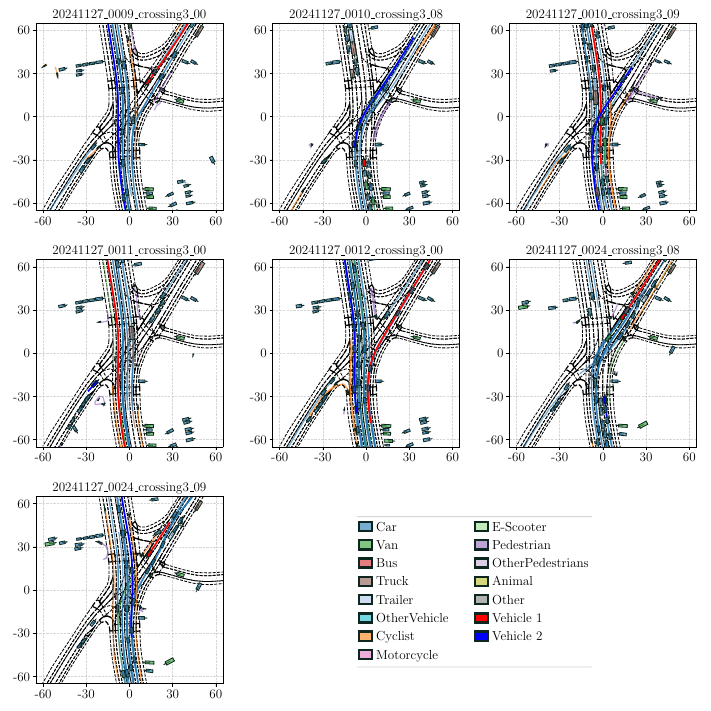}
    \caption{Visualization of trajectories at Intersection 3 across all sequences. Each subplot shows the trajectories of all annotated object classes for a sequence. The sequence names correspond to the original filenames in the dataset.}
    \label{fig:merge_tracks_c3}
\end{figure}

\clearpage
\newpage

\subsection{Intersection 1 multi-modal data visualization}\label{subsec:frame_vis_i1}

\begin{figure}[ht]
    \centering
    \includegraphics[width=\linewidth]{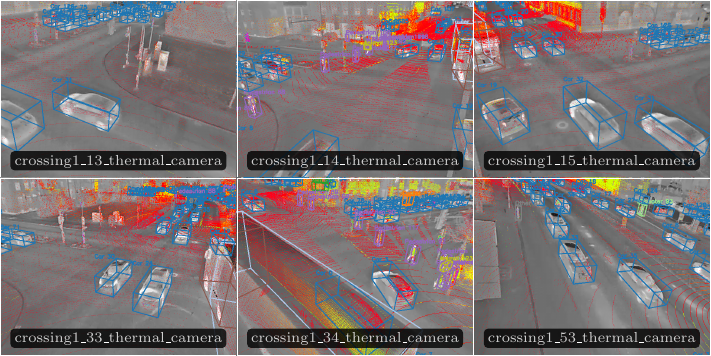}

    \vspace{1em} 

    \includegraphics[width=\linewidth]{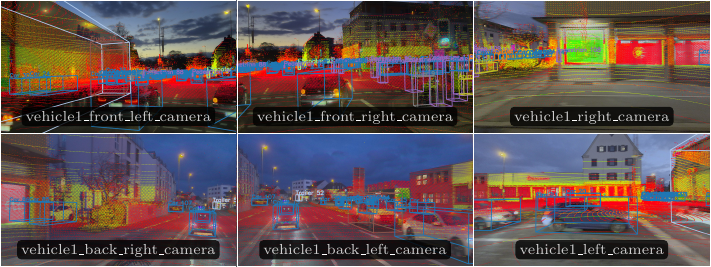}

    \vspace{1em} 

    \includegraphics[width=\linewidth]{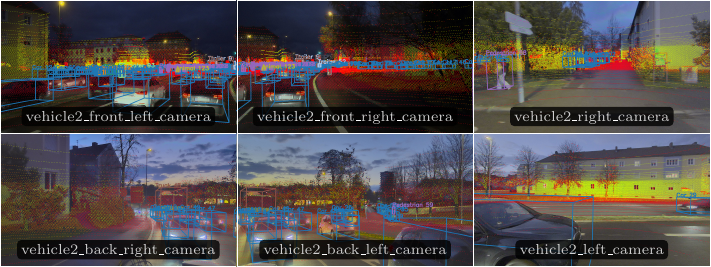}
    
    \caption{
    Multi-modal visualizations of Intersection 1 at a single timestamp, showing data from infrastructure thermal cameras (top), vehicle 1 RGB cameras (middle), and vehicle 2 RGB cameras (bottom).}
    \label{fig:crossing1_combined_views}
\end{figure}

\begin{figure}[ht]
    \centering
    \includegraphics[width=\linewidth]{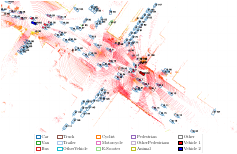}
    \caption{Visualization of the cooperative fused point cloud from all agents in Intersection 1, along with annotations.}
    \label{fig:crossing1_fused_point_cloud}
\end{figure}

\clearpage
\newpage

\subsection{Intersection 2 multi-modal data visualization}\label{subsec:frame_vis_i2}
\begin{figure}[ht]
    \centering
    \includegraphics[width=\linewidth]{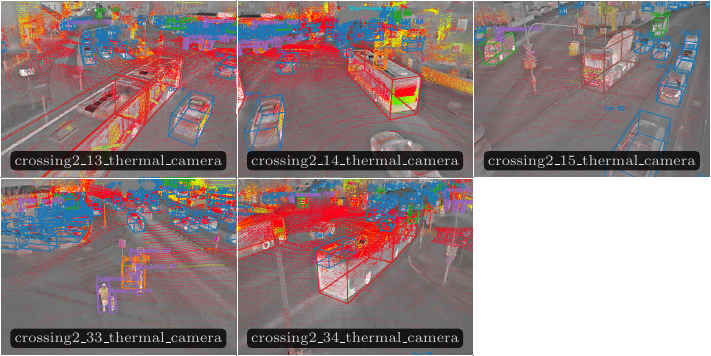}
    
    \vspace{1em} 

    \includegraphics[width=\linewidth]{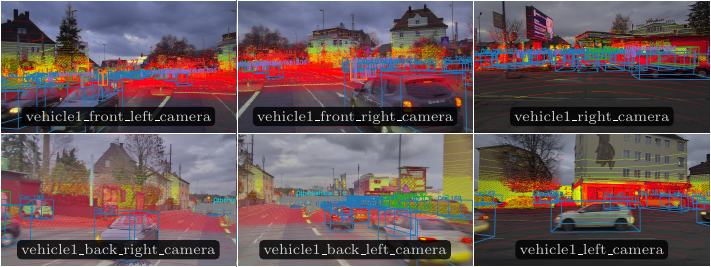}
    
    \vspace{1em} 

    \includegraphics[width=\linewidth]{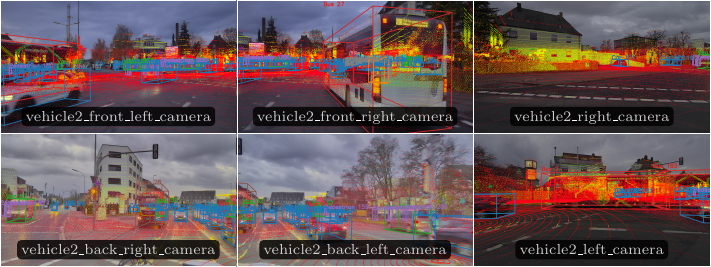}
    
    \caption{Multi-modal visualizations of Intersection 2 at a single timestamp, showing data from infrastructure thermal cameras (top), vehicle 1 RGB cameras (middle), and vehicle 2 RGB cameras (bottom).}
    \label{fig:crossing2_combined_views}
\end{figure}

\begin{figure}[ht]
    \centering
    \includegraphics[width=\linewidth]{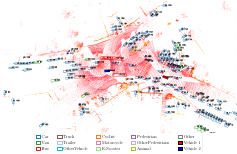}
    \caption{Visualization of the cooperative fused point cloud from all agents in Intersection 2, along with annotations.}
    \label{fig:crossing2_fused_point_cloud}
\end{figure}

\clearpage
\newpage

\subsection{Intersection 3 multi-modal data visualization}\label{subsec:frame_vis_i3}
\begin{figure}[ht]
    \centering
    \includegraphics[width=\linewidth]{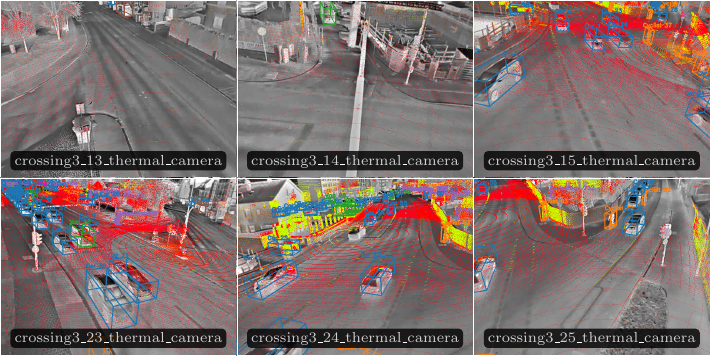}
    
    \vspace{1em} 

    \includegraphics[width=\linewidth]{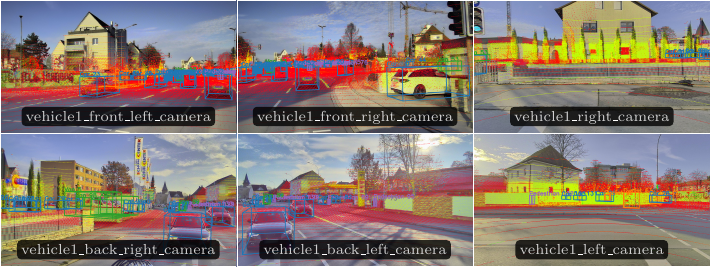}
    
    \vspace{1em} 

    \includegraphics[width=\linewidth]{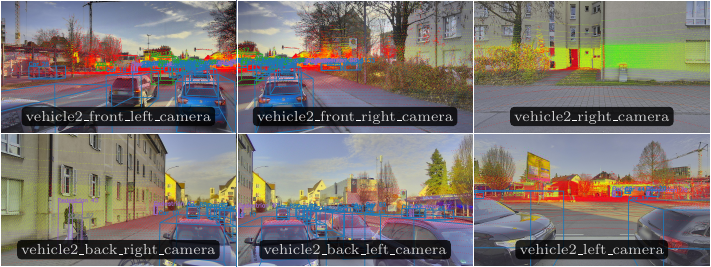}
    
    \caption{Multi-modal visualizations of Intersection 3 at a single timestamp, showing data from infrastructure thermal cameras (top), vehicle 1 RGB cameras (middle), and vehicle 2 RGB cameras (bottom).}
    \label{fig:crossing3_combined_views}
\end{figure}

\begin{figure}[ht]
    \centering
    \includegraphics[width=\linewidth]{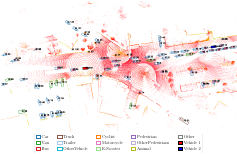}
    \caption{Visualization of the cooperative fused point cloud from all agents in Intersection 3, along with annotations.}
    \label{fig:crossing3_fused_point_cloud}
\end{figure}

\end{document}